\documentclass{article}

\PassOptionsToPackage{numbers, compress}{natbib}

\usepackage[preprint]{neurips_2025}

\usepackage{algorithm}
\usepackage{algpseudocode}
\usepackage{amsmath}
\usepackage{wrapfig}
\usepackage{pgfplots}
\usepgfplotslibrary{statistics} 
\usetikzlibrary{positioning}    
\usepackage{cutwin}
\usepackage{subcaption}
\usepackage{tabularx} 
\usepackage{array}
\usepackage{amssymb}
\usepackage{graphicx}  
\usepackage{booktabs, multirow, caption} 
\usepackage[utf8]{inputenc} 
\usepackage[T1]{fontenc}    
\usepackage[colorlinks]{hyperref}       
\usepackage{url}            
\usepackage{booktabs}       
\usepackage{amsfonts}       
\usepackage{nicefrac}       
\usepackage{microtype}      
\usepackage{xcolor}         
\usepackage{booktabs, multirow, caption} 
\usepackage{amsmath} 
\usepackage{enumitem}
\usepackage{parskip}

\usepackage{xcolor}
\usepackage{colortbl}
\usepackage{tcolorbox}
\usepackage{listings}
\definecolor{myblue}{RGB}{48, 88, 219}

\title{
\textit{AutoJudger}: An Agent-Driven Framework for \\
Efficient Benchmarking of MLLMs

\begin{cutout}{0}{-1.5cm}{0cm}{1}  
\end{cutout}
}

\author{%
    Xuanwen Ding$^{1,3}$\thanks{Equal contribution.}\,, \,
    Chengjun Pan$^{1}$\footnotemark[1]\,, \,
    Zejun Li$^1$\footnotemark[1]\,, \,
    Jiwen Zhang$^{1}$\footnotemark[1]\,, \\
    \textbf{
    Siyuan Wang$^{2}$\,, \,
    Zhongyu Wei$^{1,3}$\thanks{Corresponding author.} 
    }\\
    $^1$Fudan University, Shanghai, China \\
    $^2$University of Southern California, Los Angeles, USA \\
    $^3$Shanghai Innovation Institute, Shanghai, China \\
    \texttt{dxwpika@gmail.com} \qquad 
    \texttt{jiwenzhang21@m.fudan.edu.cn} \\
    \textcolor{myblue}{\texttt{\hyperref[https://github.com/IMNearth/AutoJudger]{https://github.com/IMNearth/AutoJudger}}}
}

\begin{document}

\maketitle

\begin{abstract}
Evaluating multimodal large language models (MLLMs) is increasingly expensive, as the growing size and cross-modality complexity of benchmarks demand significant scoring efforts. To tackle with this difficulty, we introduce \textit{\textbf{AutoJudger}}, an agent-driven framework for efficient and adaptive benchmarking of MLLMs that tackles this escalating cost. AutoJudger employs the Item Response Theory (IRT) to estimate the question difficulty and an autonomous evaluation agent to dynamically select the most informative test questions based on the model’s real-time performance. Specifically, AutoJudger incorporates two pivotal components: \textit{a semantic-aware retrieval mechanism} to ensure that selected questions cover diverse and challenging scenarios across both vision and language modalities, and \textit{a dynamic memory} that maintains contextual statistics of previously evaluated questions to guide coherent and globally informed question selection throughout the evaluation process.
Extensive experiments on four representative multimodal benchmarks demonstrate that our adaptive framework dramatically reduces evaluation expenses, i.e. AutoJudger uses only 4\% of the data to achieve over 90\% ranking accuracy with the full benchmark evaluation on MMT-Bench. 
\end{abstract}

\section{Introduction}
\label{section:Intro}
\begin{figure}[t]
  \centering
  \setlength{\abovecaptionskip}{1pt}
  \setlength{\belowcaptionskip}{0pt}
  \begin{subfigure}{0.455\textwidth}
    \centering
    \setlength{\abovecaptionskip}{0pt}
    \includegraphics[width=\textwidth]{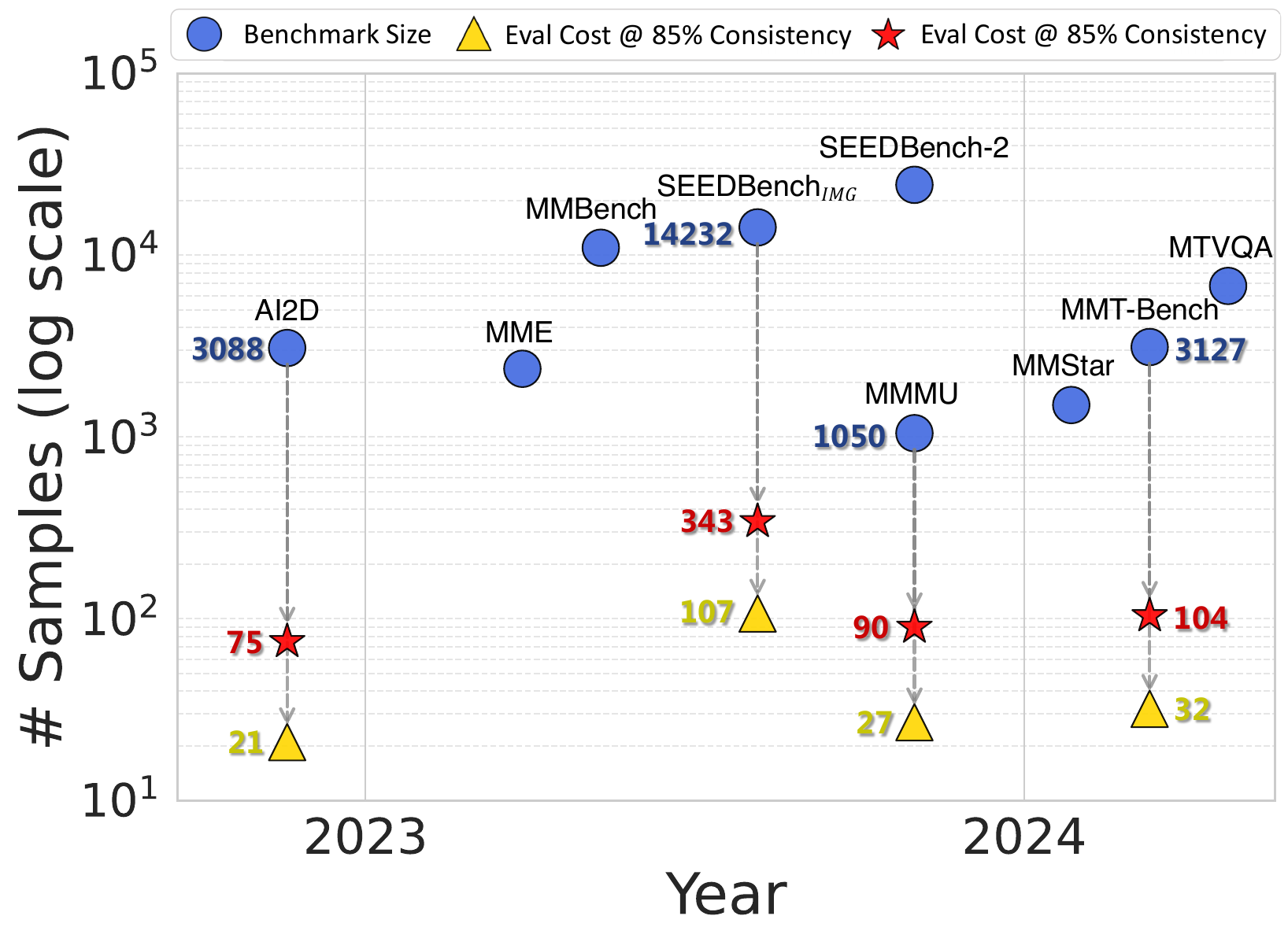}
    \caption{Number of samples in benchmarks over time. }
    \label{fig:various_benchmarks}
  \end{subfigure}
  \hfill
  \begin{subfigure}{0.455\textwidth}
    \centering
    \setlength{\abovecaptionskip}{0pt}
    \includegraphics[width=\textwidth]{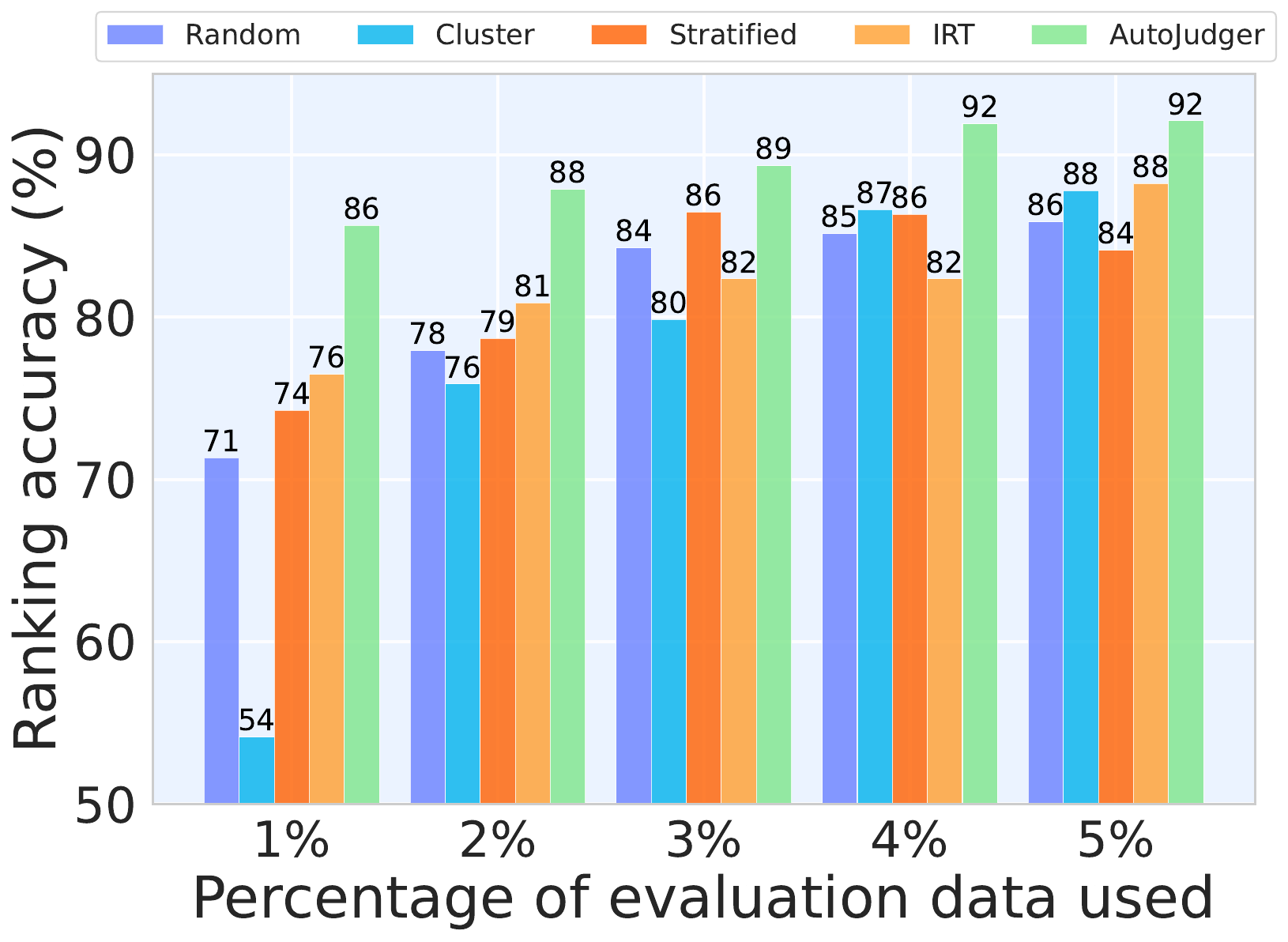}
    \caption{Ranking consistency on MMT-Bench.}
    \label{fig:another}
  \end{subfigure}
  \caption{\textbf{Benchmark scale and efficiency of AutoJudger.}
    (a) plots the scales of various benchmarks that are commonly adopted in MLLM evaluation. The triangle and pentagon markers indicate the number of samples required by AutoJudger to achieve 85\% and 90\% consistency with the full-set evaluation results, respectively. 
    (b) compares several efficient benchmarking methods on MMT-Bench. AutoJudger achieves 92\% rank consistency using only 4\% of the data (125 samples).}
  \label{fig:comparison}
\end{figure}

Motivated by the success of Large Language Models (LLMs)~\cite{achiam2023gpt,touvron2023llama,yang2024qwen2,liu2024deepseek}, Multimodal Large Language Models (MLLMs)~\cite{hurst2024gpt,liu2023visual,bai2025qwen2,chen2025janus} have been developed to tackle challenging tasks involving the joint understanding and generation of information across multiple modalities, such as text and images~\cite{li2024continuous,li2025survey}.
To assess the full spectrum of MLLM capabilities, a growing number of benchmarks have been introduced as illustrated in Figure~\ref{fig:various_benchmarks}, spanning diverse domains~\cite{fu2023mme,liu2024mmbench,fu2024blink,yue2024mmmu,li2023seed,li2024reform}. However, this also introduces a computational burden for comprehensive capability evaluation.

Compared to text-only scenarios, the evaluation cost problem becomes more pronounced in multimodal benchmarks, as including visual contexts substantially lengthens the input sequences~\cite{terragni2024evaluating,xu2025collageprompt}.
In addition, incorporating reasoning-enhancement methods like chain-of-thought~\cite{wei2022chain,guo2025deepseek} and employing ChatGPT to assist in scoring model responses~\cite{liu2024mmbench,lu2023mathvista} will further increase the cost.
This raises a critical question: \textbf{Can we evaluate MLLMs more efficiently without sacrificing reliability?}

To address this problem, a line of studies focuses on exploring efficient benchmarking methods~\cite{perlitz2023efficient,polo2024tinybenchmarks,vivek2023anchor}: selecting a subset from the benchmark for efficient evaluation while maintaining consistency with the results obtained from full-set evaluation. 
Existing approaches are primarily designed for text-only benchmarks, performing stratified sampling based on question categories~\cite{perlitz2023efficient} and difficulty levels~\cite{zhuang2023static,polo2024tinybenchmarks} to construct subsets for evaluation.
However, multimodal scenarios pose additional challenges: (i) Most multimodal benchmarks do not explicitly assess or characterize question difficulty; (ii) Each image-question pair contains rich multimodal semantic information, merely relying on coarse-grained information like question categories struggle to ensure semantic diversity within the subset; (iii) There exists large performance variance across models. Assigning the same subset of questions to all models may limit the efficiency in distinguishing between models, e.g., evaluating powerful models with too many simple questions provides little information gain.

Facing above challenges, an ideal evaluation system needs to comprehensively consider factors including the performance of evaluated models, question difficulties and semantics, iteratively constructing subsets during assessment. To tackle such a dynamic decision-making problem, we propose \textbf{\textit{AutoJudger}}, an agent-driven multimodal evaluation framework. We formulate efficient benchmarking as an interview scenario, where an agent powered by MLLM serves as the interviewer, continuously interacting with the environment (the question pool and the evaluated models) to select appropriate questions, dynamically assessing model capabilities throughout the interview.

Furthermore, we design three modules to assist AutoJudger in the interaction with the interview environment and the evaluation process. (i) We collect extensive offline evaluation results of various MLLMs and characterize the difficulty of benchmark questions based on Item Response Theory (IRT)~\cite{cai2016item}. This difficulty framework supports subsequent question selection and real-time performance assessment during the interview.
(ii) Considering the large scale of the question pool, we augment AutoJudger with a multimodal semantic-aware retrieval module to access the entire benchmark. The retrieval module performs a coarse filtering process, the interviewer agent then conduct analysis and selection on the retrieved candidates. This strategy fully leverages sample-level semantics to ensure richness of the selected subset while maintaining efficiency.
(iii) We introduce a dynamic memory module to help the agent summarize information about previously tested questions and the model performance. This module assists the agent in making personalized question selections for different models during the evaluation process and provides an interpretable analysis of model capabilities.

AutoJudger can be seamlessly integrated into the evaluation of any MLLMs in a plug-and-play manner. 
We conduct extensive experiments to demonstrate that AutoJudger significantly reduces evaluation costs across different benchmarks while maintaining the reliability and stability.
Despite the growing size of modern benchmarks, AutoJudger still achieves remarkable rank consistency—85\% and even 90\%—with only a small fraction of the samples, as shown in Figure~\ref{fig:comparison}.
For instance, on MMT-Bench, AutoJudger reaches 92\% rank consistency using merely 4\% of the data (125 samples).

The key contributions are summarized as follows:
\begin{itemize}[topsep=1pt,leftmargin=*]
\setlength{\itemsep}{3pt}
\setlength{\parsep}{2pt}
\setlength{\parskip}{0pt}  
    \item We propose \textit{\textbf{AutoJudger}}, the first agent-driven framework for efficient benchmarking of MLLMs. Unlike prior static methods, AutoJudger adaptively selects informative questions by interacting with evaluated models, leveraging the reasoning ability of the judging agent to guide the evaluation. 
    \item To jointly capture the question difficulty and the cross-modal semantic diversity, we equip AutoJudger with a semantic-aware retrieval mechanism supported by Item Response Theory (IRT), ensuring that selected questions are representative. To further enhance the adaptivity, we incorporate a dynamic memory that maintains contextual statistics of previously evaluated questions, enabling coherent and globally informed question selection throughout the evaluation process.
    \item We conduct extensive experiments on four commonly-adopted multimodal benchmarks to evaluate the performance of 17 MLLMs with AutoJudger. The results demonstrate that our adaptive framework significantly reduces evaluation costs—for example, AutoJudger achieves over 90\% ranking accuracy on MMT-Bench using only 4\% of the full benchmark data.
\end{itemize}

\section{Related Work}

\subsection{Multi-modal Benchmarks} 
As MLLMs continue to evolve, a series of large-scale benchmarks have been proposed to assess their capabilities across diverse scnarios and  tasks~\cite{xu2024lvlm,liu2024ocrbench,lu2023mathvista,lu2022learn}.
Early efforts such as AI2D~\cite{kembhavi2016diagram} focus on diagram-based reasoning, providing over 5,000 diagrams and 15,000 QA pairs.
MMMU~\cite{yue2024mmmu} targets expert-level subject reasoning with 11,500 questions across six disciplines and 183 subfields.
CMMU~\cite{he2024cmmu} and CMMMU~\cite{ge2024cmmmu} further extend this work to Chinese.
MMT-Bench~\cite{ying2024mmt} scales the scope to 31K+ questions spanning 32 meta-tasks and 162 subtasks, covering settings like autonomous driving and embodied navigation.
SEED-Bench~\cite{li2023seed} broadens the evaluation space with 19K questions across 12 dimensions, later expanded to 24K and 27 dimensions in SEED-Bench-2~\cite{li2024seed}.
While these benchmarks provide comprehensive coverage, the large scale makes evaluation increasingly expensive.
In addition, some benchmarks employ proprietary models like ChatGPT to assist in scoring model response~\cite{liu2023visual,liu2024mmbench,lu2023mathvista}, leading to a rapid increase in evaluation costs.

\subsection{Efficient Benchmarking} 

To reduce the rising cost of large-scale evaluations, recent studies have focused on efficient benchmarking—selecting smaller, informative subsets that preserve evaluation fidelity.
Broadly, existing approaches fall into two categories: feature-based sampling and difficulty-based sampling.
Feature-based sampling aim to reduce redundancy by selecting semantically representative examples.
For instance, Perlitz et al.~\cite{perlitz2023efficient} perform stratified sampling based on task sub-scenarios to ensure diversity, while Anchor Points~\cite{vivek2023anchor} group examples by model confidence to prioritize high-impact questions.

The other line of methods sample subsets based on difficult levels of questions.
Zhuang et al.~\cite{zhuang2023efficiently} leverage IRT~\cite{cai2016item} to tailor questions to each model. TinyBenchmark~\cite{polo2024tinybenchmarks} clusters questions by difficulty to construct subsets.
Easy2Hard-Bench~\cite{ding2024easy2hard} combines IRT with Glicko-2~\cite{glickman2012example} to estimate difficulty from historical performance.
Beyond specific perspectives, \textit{\textbf{AutoJudger}} unifies both semantic diversity and difficulty adaptiveness to allow efficient and adaptive benchmarking of MLLMs.

\subsection{Difficulty Estimation} 
Most existing benchmarks for LLMs and MLLMs lack explicit difficulty annotations, limiting the granularity of capability evaluation. Classical psychometric frameworks such as IRT\cite{cai2016item} model the probability of a correct response as a function of latent question difficulty and subject ability. Traditionally used in standardized exams like the GRE and SAT\cite{an2014item}, IRT has been extended to NLP settings to analyze benchmark saturation~\cite{vania2021comparing} and estimate model proficiency~\cite{park2024large}. These applications demonstrate IRT’s flexibility in both question-level diagnostics and model ranking, making it a natural fit for difficulty estimation in LLM evaluation. Other lines of work approach difficulty estimation from content-based features~\cite{jiao2023automatic,xu2024adaption}, step-level reasoning complexity~\cite{cheng2021guiding,wang2024benchmark}, or LLM-based prediction models~\cite{gao2018difficulty,lee2023difficulty}. 
In our work, we incorporate IRT into the adaptive benchmark construction process, where question difficulty is pre-estimated from historical model responses and then used to enable dynamic, targeted evaluations across varying model capabilities.

\section{Problem Statement of Efficient Benchmarking}
\label{section:problem}

Efficient benchmarking aims to reliably evaluate a model on a specific benchmark with as less expenses as possible. In this work, we constrain the scope of evaluated models to MLLMs, due to the rapidly increasing cost of these models and the lack of prior work in this area. 
We denote the complete evaluation benchmark as $Q=\{q_i\}_{i=1}^{N}$ where each $q_i$ is a test question, representing a multimodal question-answer pair. Given a candidate model $m$, the objective is to find a mapping $f: Q \rightarrow \hat{Q}$ such that the performance of model $m$ on the selected subset $\hat{Q}$, denoted as $P(m|\hat{Q})$, is consistent with its performance on the full benchmark set $Q$, i.e.
\begin{equation}
    \rho\left( P(m|\hat{Q}), P(m|Q) \right) \ge \sigma
\end{equation}
where $\rho$ represents the consistency scoring function, and $\sigma$ is the consistency threshold. 
Typically, the consistency is estimated by comparing the ranking of a evaluated model $m_j$ on a group of models $M=\{m_j\}$. 
Therefore, the problem of efficient benchmarking is formulated as follows:
\begin{equation}
    \max_{f: Q \rightarrow \hat{Q}}  \rho\left( 
    \{ P(m_j|\hat{Q})\}_{m_j \in M }, 
    \{ P(m_j|Q)\}_{m_j \in M } \right)  \quad
    \text{s.t.} \quad \hat{Q}\subset Q, \, |\hat{Q}|\leq \delta * |Q|
\end{equation}
where $\delta$ denotes the compression ratio. In this work, $\delta$ is set to 5\% unless otherwise specified.

\section{\textit{AutoJudger}}
\label{section:method}

\begin{figure}[t]
  \centering
  \setlength{\abovecaptionskip}{0pt}
  \includegraphics[width=1.0\linewidth]{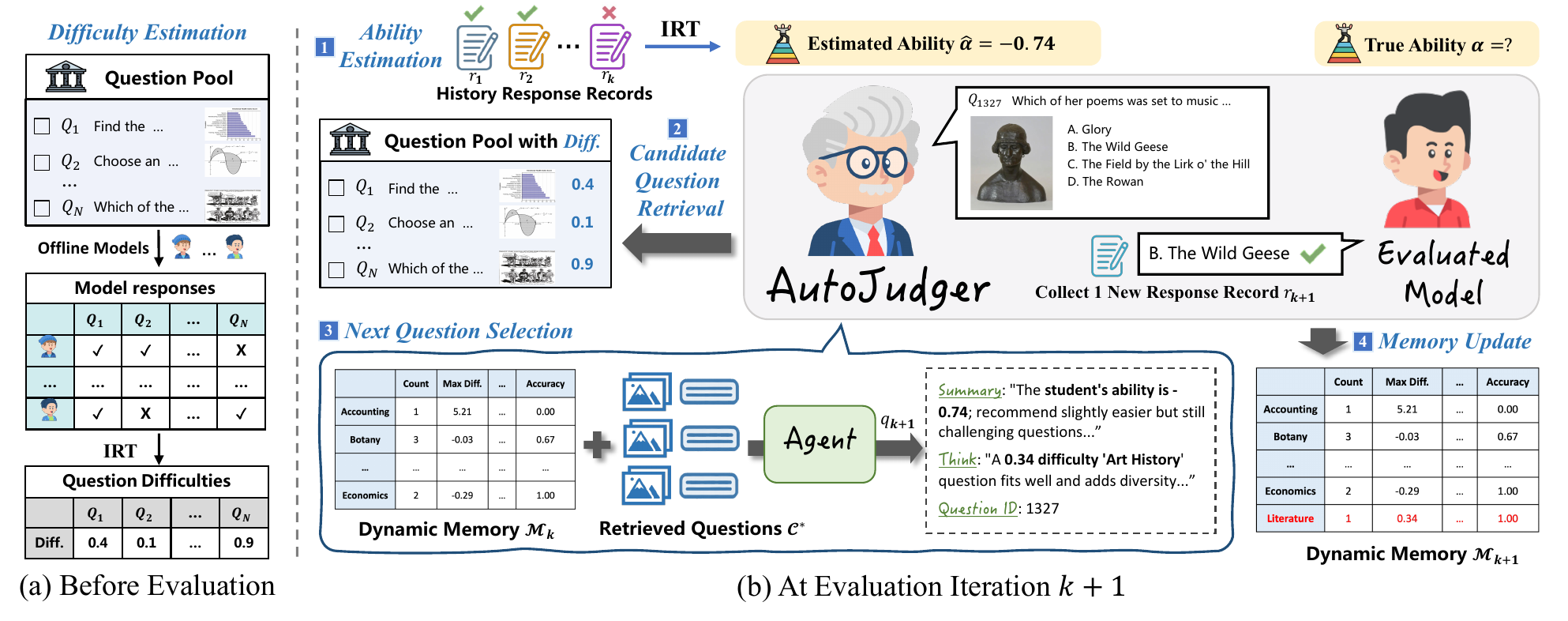}
  \caption{\textbf{The framework of AutoJudger.} 
  Before evaluation, the difficulties of question from a benchmark are computed by utilizing a set of offline models. At each evaluation iteration, AutoJudger firstly retrieve the candidate questions based on the estimated ability. Then, AutoJudger selects the most proper question, collect the response from the evaluated model, and update its memory. }
  \label{figure:framework}
\end{figure}

To address this problem, we propose \textit{\textbf{AutoJudger}}, an agent-driven efficient benchmarking framework as shown in Figure~\ref{figure:framework}. Our underlying intuition is that, analogous to human interview, the effective evaluation of a model entails dynamically selecting the most appropriate questions based on its real-time performance. Therefore, we formulate the construction of mapping $f$ as as a dynamic decision-making problem. Specifically, the next question used to evaluate model $m_j$ is selected by considering the memory $\mathcal{M}_k$ of previously attempted questions $Q'_k=\{q_i\}_{i=1}^{k}$ and responses $\{r_{ij}\}_{i=1}^{k}$, current model performance $P\left(m_j|\{q_i\}_{i=1}^{k}\right)$ and the complete evaluation benchmark $Q$:
\begin{equation}
    q_{k+1} = f\left( \mathcal{M}_k, P\left(m_j|Q'_k\right), Q \right), \quad 
    \mathcal{M}_k=\{q_i, r_{ij}\}_{i=1}^{k}
\end{equation}
To select appropriate samples, AutoJudger should acquire a clear understanding of the difficulty of questions during evaluation. To this end, we first leverage Item Response Theory (IRT)~\cite{cai2016item} to characterize the difficulty $d_i$ of each question $q_i$ before the evaluation (\S~\ref{subsection:IRT}). We then employ an intelligent agent powered by an MLLM to recommend questions of appropriate difficulty. Such recommendation is based on the current estimation of model's ability (\S~\ref{subsection:ability_estimation}). Since the full benchmark $Q$ is too large to be processed efficiently, we design a \textit{semantic-aware retrieval mechanism} to reduce the candidate pool size (\S~\ref{subsection:retrieval}). This mechanism also ensures that the selected questions span diverse and challenging scenarios. Subsequently, we prompt an MLLM-based agent to choose the most appropriate question from the retrieved candidate questions (\S~\ref{subsection:agent}). After the evaluated model answers a newly selected question, we record its response. The history of attempted questions and model responses is summarized by the agent, stored and updated in the \textit{memory} $\mathcal{M}$ of the agent to guide coherent and globally informed question selection during evaluation (\S~\ref{subsection:memory}).

\subsection{Question Difficulty Estimation}
\label{subsection:IRT}

To preform efficient sample selection, AutoJudger adopts Item Response Theory (IRT)~\cite{cai2016item} to estimate the difficulty of the questions from a specific benchmark by utilizing a set of offline MLLMs before evaluation. Note that, to prevent information leakage, we carefully ensure that there is no overlap between the offline MLLMs and the models under evaluation (see Appendix~\ref{appendix:implementation} for details).

\paragraph{Modeling with IRT}  
 We adopt a one-parameter logistic IRT model, also known as the Rasch model~\cite{rasch1993probabilistic}, which defines the correct probability of a response $r_{ij}$ of model $m_j$ on question $q_i$ as:
\begin{equation}
    p\left(r_{ij} \, \text{is correct}\right) = \frac{1}{1 + \exp(-(a_{j} - d_{i}))}
    \label{eq:1PL}
\end{equation}
where $a_{j}$ is the latent ability of the model $m_j$ , and $d_i$ is the difficulty of the question $q_i$. Intuitively, a model is more likely to succeed on questions with lower difficulty than its ability level.

\paragraph{Estimating Question Difficulty}  
Given a collection of response records $\{r_{ij}\}$ from a set of previously evaluated offline models $M'=\{m'_j\}$ on benchmark $Q=\{q_i\}$, we fit the Rasch model to estimate the question difficulties $D=\{d_i\}$ via maximum likelihood estimation, using the Bayesian variational framework proposed by \cite{ding2024easy2hard} (please refer to Appendix~\ref{appendix:rasch_fitting} for more details). Once estimated, these difficulty scores are fixed and serve as priors for subsequent question selection.

\subsection{Model Ability Estimation}
\label{subsection:ability_estimation}
Given the estimated question difficulties $D=\{d_i\}$, IRT enables the real-time assessment of model ability based on its response history, supervising the model proficiency during evaluation.
When evaluating the model $m_j$ at $k$-th iteration, according to Equation~\eqref{eq:1PL}, we use maximum likelihood estimation to infer the current ability $a_{j,k}$ based on the selected questions $Q'_k$, model responses $\{r_{ij}\}_{i=1}^k$, and the fixed difficulties $D$. 
We develop a binary search algorithm to efficiently find the optimum for $a_{j,k}$, as detailed in Appendix~\ref{appendix:abi_estimation}. 
This allows us to track the model’s progress over time and guide the adaptive selection of future questions. Notably, we define the model performance $P(m_j|Q'_k)$ as its current ability value $a_{j,k}$ estimated through IRT. 

\subsection{Candidate Question Retrieval}
\label{subsection:retrieval}

Given the large scale of existing benchmarks, it is impractical for the agent to directly select questions from the entire question pool $Q$. To address this challenge, we design a retrieval strategy to provide the agent with a candidate set $\mathcal{C}^*_k$ with feasible size $|\mathcal{C}^*_k| \ll |Q|$. We aim to select questions that are both appropriate in difficulty and semantically distinct from those previously attempted in $Q'_k$.
\begin{equation}
\label{eq:judgerfunc}
    q_{k+1} = f\left( \mathcal{M}_k, a_j, \mathcal{C}^*_k \right), \quad \mathcal{C}^*_k\subset Q 
\end{equation}

\paragraph{Initialization} 
At the beginning, the ability of the model $P(m_j,\varnothing)$ remains unknown. To build a strong starting point, we adopt a clustering-based strategy to ensure semantic diversity across the initial set $Q'_0$. We first encode each question with a semantic feature extractor (e.g., CLIP~\cite{radford2021learning}, Qwen2.5-VL~\cite{bai2025qwen2}) to obtain meaningful embeddings. Based on these embeddings, we perform K-means clustering over the entire benchmark to group questions with similar semantics.
From each cluster, we uniformly sample a small number of questions to construct the initial candidate pool $Q'_0$. This approach ensures broad semantic coverage and avoids the early-stage selection bias, enabling robust ability estimation and question selection in later adaptive steps.

\paragraph{During Iteration} 
Firstly, we filter out questions that are either too hard or too easy for the current model. We compute the estimated probability $p$ of the target model $m_j$ correctly answering each question $q_i\in Q/Q'_k$, based on its current ability $a_{j}$ and the question difficulty $d_{i}$, using the IRT formulation defined in Equation~\eqref{eq:1PL}. We keep a candidate set $\mathcal{C}_k$ to retain only questions whose success probabilities fall within a desirable range (we set $p_{\text{min}}$ as 0.2 and $p_{\text{max}}$ as 0.8):
\begin{equation}
    \mathcal{C}_k = \{ i \in Q/Q'_k \mid p_{\min} \leq p \leq p_{\max} \}
    \label{eq:cand_subset}
\end{equation}
Subsequently, to encourage semantic diversity, we apply a max-min retrieval strategy. For each candidate $q \in \mathcal{C}_k$, we compute its distance to the previously selected question set $Q'_k$ and select the questions with the maximum distance:
\begin{equation}
    \mathcal{C}^*_k = \left\{q_1^*, ..., q_5^* \,|\, q^* = \arg\max_{q \in \mathcal{C}_k} \min_{q' \in Q'_k} \text{dist}(q, q') \right\}
    \label{eq:cand_num}
\end{equation}
We retain the top-5 questions that exhibit the greatest Euclidean distance from $Q'_k$ as the final candidate set $\mathcal{C}^*_k$. Consequently, such a semantic-aware retrieval strategy ensures that the selected questions not only align with real-time ability but also introduce novel semantic coverage.

\subsection{Next Question Selection}
\label{subsection:agent}

Based on the retrieved results $C^*_k$, we prompt the interviewer agent to perform fine-grained analysis on candidates questions and recommend the next question for the tested model as: 
\begin{equation}
    q_{k+1} = f_{\theta}(\mathcal{M}_k, a_{j,k}, C^*_k,D^*_k)\;\;\;\; \text{where}\;\;q_{k+1}\in C^*_k
\end{equation}
As the agent is powered by strong MLLM $f_{\theta}$, we leverage its multimodal understanding capability to analyze each question, providing its reasoning process by comprehensively considering previous interview history $\mathcal{M}_k$, the real-time model capability $a_{j,k}$, the detailed semantics of candidate questions $C^*_k$ and the corresponding difficulties of candidate questions $D^*_k$.
Ultimately, the agent selects the most appropriate question $q_{k+1}$ from these five candidates to serve as the next evaluation question, and we update the previously attempted question set as $Q'_{k+1} \leftarrow Q'_k \cup \left\{q_{k+1}\right\}$. Detailed prompts used for the interviewer agent are provided in Appendix~\ref{appendix:prompt}.

\subsection{Dynamic Memory Update}
\label{subsection:memory}

To maintain contextual coherence during evaluation, AutoJudger supports the memory mechanism. Emphasizing long-term statistical awareness, the memory $\mathcal{M}$ accumulates high-level information about previously selected questions and model responses, grouped by semantically inferred categories as a markdown table. Since many benchmarks lack predefined class labels or contain noisy annotations, AutoJudger assigns categories based on semantic features and dynamically expands the category table as new topics emerge. For each category, the memory tracks statistics including the number of questions, max/min/average difficulty, and overall accuracy. This enables the agent to maintain global awareness of coverage and balance across domains. A representative example is:
\begin{table}[h]
\setlength{\belowcaptionskip}{0pt}
\setlength{\abovecaptionskip}{0pt}
\small
\centering
\begin{tabular}{l|c|c|c|c|c}
\toprule
\textbf{Category} & \textbf{Count} & \textbf{Max Difficulty} & \textbf{Min Difficulty} & \textbf{Avg Difficulty} & \textbf{Accuracy} \\
\midrule
Accounting & 5 & 5.21 & -1.02 & 1.15 & 0.60 \\
Art History & 20 & 1.01 & -5.20 & -0.83 & 0.71 \\
Botany & 9 & 0.45 & -5.30 & -1.31 & 0.56 \\
Cell Biology & 14 & 4.90 & -2.44 & -0.70 & 0.50 \\
\bottomrule
\end{tabular}
\end{table}

This memory table illustrates a more realistic usage scenario, featuring a broad range of question difficulties and imbalanced category distributions. For example, Art History contains 20 questions spanning a wide difficulty range with relatively high accuracy, while Accounting tend to be more difficult with greater outcome variance. Such statistical tracking allows the agent to identify underrepresented or overly challenging areas, informing more targeted selection in subsequent iterations.

\section{Experiment}
\subsection{Experiment Setups} 
\label{subsection:setup}

\paragraph{Benchmarks} We validate the effectiveness of AutoJudger on four commonly-adopted benchmarks: MMMU-Dev Val~\cite{yue2024mmmu}, SEEDBench-Image~\cite{li2024seed}, MMT-Bench-Val~\cite{ying2024mmt}, and AI2D-Test~\cite{kembhavi2016diagram}. AI2D represents a relatively simple scenario, while the other three benchmarks are used for comprehensive multi-dimensional evaluation, simulating complex environment with a diverse question pool.

\paragraph{Metrics}
We propose a metric, ranking accuracy, to quantitatively assess the consistency between our evaluation results and the results on the full benchmark, defined as:
\begin{equation}
\rho = \text{Ranking Accuracy}(\%)=\left( 1-\frac{\text{\# Inversions}}{n*(n-1)/2} \right)*100
\end{equation}
where the number of inversions refers to pairwise discrepancies between the predicted and ground-truth rankings, where the ground-truth ranking is determined by the model accuracy over the full benchmark.
In addition to consistency, efficient benchmarking methods are supposed to be stable. We also report the confidence intervals (1.96 times the standard deviation) of the ranking accuracy based on multiple experiments. A narrower interval indicates the corresponding method is more stable.

\paragraph{Baselines} 
We compare AutoJudger against a set of baseline strategies categorized into two groups: unified sampling and model-specific sampling. Unified sampling methods use a shared question pool for all models, including \textit{Random Sampling (Random)}, which selects a fixed number of questions at random; Stratified Random Sampling (Stratified)~\cite{perlitz2023efficient} also selects a fixed number of questions randomly, but applies weighted sampling based on the number of categories within each benchmark; and \textit{Cluster-Based Sampling (Cluster)}, which applies K-means clustering to the BERT embeddings of questions and selects those closest to each cluster centroid as the evaluation questions. In contrast, model-specific sampling strategies adapt question selection to individual models. The most representative method is \textit{Optimal IRT Difficulty Choosing (IRT)}~\cite{lord2012applications}, which iteratively selects questions whose IRT-estimated difficulty is closest to the model's latent ability until the desired sample size is reached. Appendix~\ref{appendix:implementation} provides a detailed introduction to baseline methods.

\paragraph{Implementation Details}
We deploy our AutoJudger framework based on the Qwen2.5-VL-7B-Instruct model~\cite{bai2025qwen2} where the retrieval module is driven by CLIP ViT-B/32~\cite{radford2021learning}. 
For IRT-based difficulty assessment of the questions, we collect offline evaluation results from 60 models (training set)\footnote{Offline results are collected by VLMEvalKit~\cite{duan2024vlmevalkit}: \hyperref[https://github.com/open-compass/VLMEvalKit]{\texttt{https://github.com/open-compass/VLMEvalKit}}}. During evaluation, we used AutoJudger to assess another 17 models (test set). Both subsets of models cover a wide range of parameter scales, including both open-source and proprietary models, with no overlap between them. Please refer to Appendix~\ref{appendix:implementation} for more details. Each experiment is repeated five times to reduce the impact of randomness. 
We conduct all experiments on a Linux machine running Ubuntu 22, with 8 NVIDIA RTX 4090 GPUs.

\begin{table}[t]
\small
\centering
\caption{\textbf{Performance of different methods under 5\% compression ratio}. We report the average ranking accuracy together with the confidence intervals. The best results are highlighted in \textbf{bold}.}
\label{tab:baseline_performance}
\begin{tabular}{c|cccc}  
\toprule
\textbf{Method} & \textbf{AI2D}$_{\textit{TEST}}$ & \textbf{MMMU}$_{\textit{DEV VAL}}$ & \textbf{MMT-Bench} & \textbf{SEEDBench}$_{\textit{IMG}}$    \\
\midrule
Random & 93.82±3.71 & 81.47±6.28 & 85.88±6.14 & \textbf{92.65±4.74} \\
Cluster & 93.53±3.80 & 78.97±11.62 & 87.79±3.71 & 92.50±3.58 \\
Stratified & 93.97±3.08 & 84.26±6.22 & 84.12±6.61 & 90.88±3.82 \\
IRT & 89.71±0.00 & 82.35±0.00 & 88.24±0.00 & 91.91±0.00 \\
\midrule
\textit{\textbf{AutoJudger}} & \textbf{94.85±0.00} & \textbf{87.94±0.71} & \textbf{92.06±1.41} & 90.74±0.71 \\
\bottomrule
\end{tabular}
\end{table}

\begin{figure}[t]
  \centering
  \setlength{\abovecaptionskip}{1pt}
  \begin{subfigure}{\textwidth}
    \centering
    \setlength{\abovecaptionskip}{0pt}
    \includegraphics[width=\textwidth]{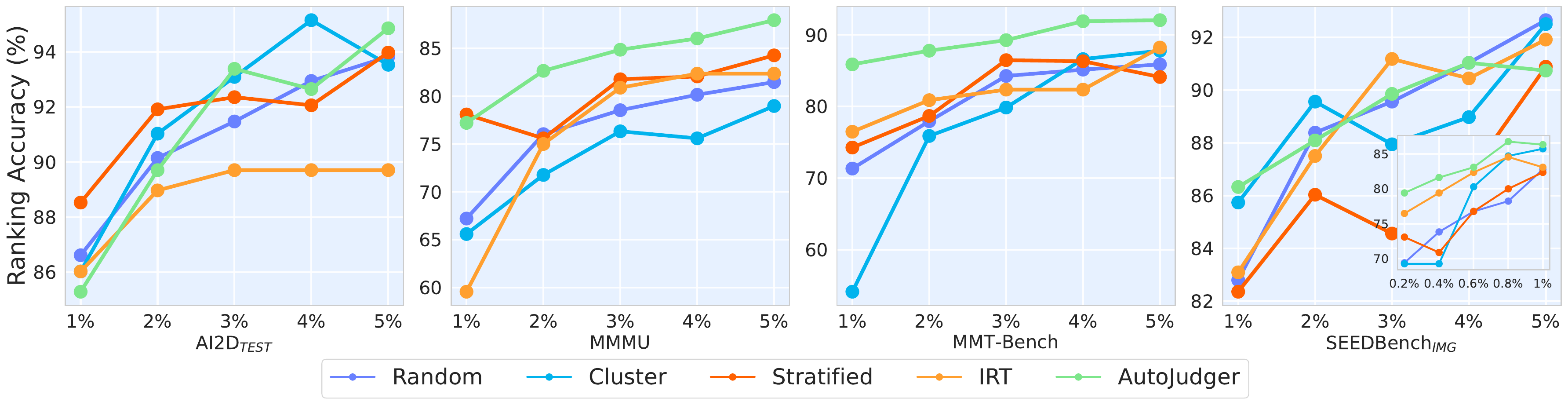}
    \caption{Average ranking accuracy.}
    \label{fig:main_linchart}
  \end{subfigure}
  \\
  \begin{subfigure}{\textwidth}
    \centering
    \setlength{\abovecaptionskip}{0pt}
    \includegraphics[width=\textwidth]{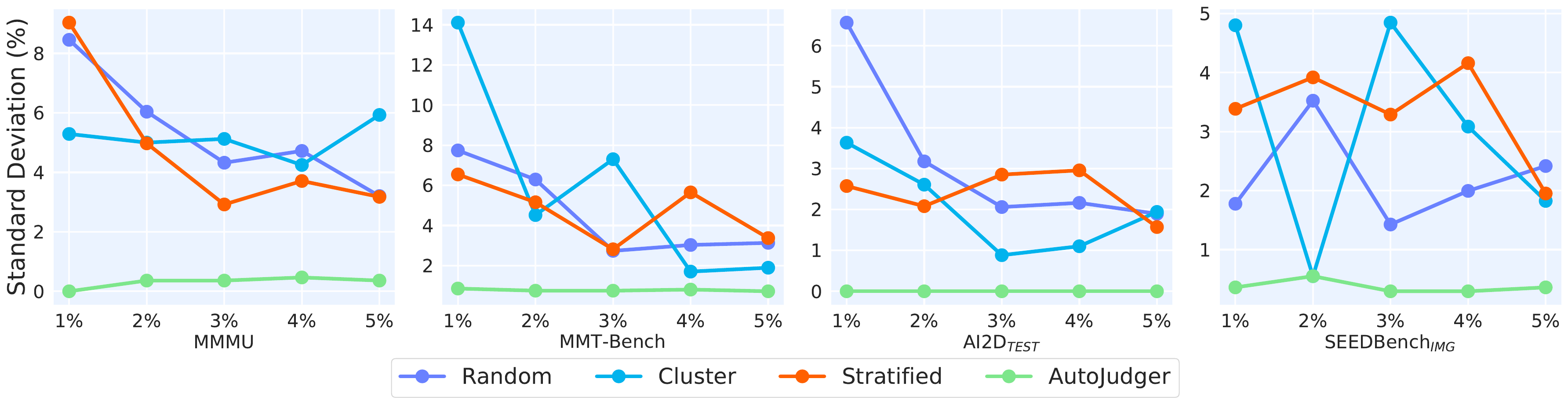}
    \caption{Standard deviation of ranking accuracy.}
    \label{fig:std_linchart}
  \end{subfigure}
  \caption{\textbf{Evaluation performance and stability under varying compression ratios.}}
  \label{fig:lincharts_combined}
\end{figure}
\subsection{Main Results}

In Table~\ref{tab:baseline_performance}, we compare our framework with several efficient benchmarking baselines using only 5\% of the data across four widely-used benchmarks. Three key observations emerge:
(1) AutoJudger consistently outperforms all baselines on most benchmarks, demonstrating strong effectiveness in low-data regimes.
(2) Compared to stochastic baselines (e.g., Random, Cluster, Stratified), AutoJudger exhibits significantly lower variance, indicating enhanced stability and robustness.
(3) The IRT baseline, being a deterministic method, does not introduce randomness, but it suffers from suboptimal performance--for example, IRT performs notably worse on MMMU.
In contrast, AutoJudger integrates real-time model feedback and history records to dynamically select appropriate questions. This adaptive strategy gradually mitigates the negative effects of randomness, improving the stability while achieving superior overall performance.

Furthermore, in Figure~\ref{fig:lincharts_combined}, we compare the performance and stability of different methods under varying compression ratios. (1) In terms of evaluation accuracy, all baselines perform well on AI2D, a relatively simple benchmark. However, on complex benchmarks like MMT and MMMU, AutoJudger demonstrates consistent and significant advantages.
We believe the reason why AutoJudger shows less improvement on SEEDBench is that the dataset itself is relatively large—about four times the size of the others—so even 5\% already includes a substantial amount of data for baselines to converge. Therefore, we further explore smaller compression rates and find that the advantage of AutoJudger becomes very prominent below a 1\% compression rate. In summary, these results demonstrate that AutoJudger can achieve reliable evaluation on complex benchmarks at a much lower cost.

(2) From the perspective of stability, it is obvious that all baseline methods require a large subset to ensure stable evaluation, whereas--as discussed earlier--AutoJudger applies an adaptive strategy which effectively maintains stability across different data scales.

\begin{table}[t]
\small
\centering
\caption{\textbf{Impact of different framework design of AutoJudger on four benchmarks.}}
\label{tab:Ablation Studies}
\begin{tabular}{lcccc}
\toprule
\textbf{Method Variant} 
&{$\text{AI2D}_{\textit{TEST}}$} & \textbf{MMMU}$_{\textit{DEV VAL}}$ & \textbf{MMT-Bench} & \textbf{SEEDBench}$_{\textit{IMG}}$    \\
\midrule
\textit{\textbf{AutoJudger}} & \textbf{94.85} & 88.24 & \textbf{93.38} & \textbf{91.18} \\
$w/o$ agent        & 92.28 & 82.87 & 86.62 & 91.10 \\ 
$w/o$ visual       & 94.85 & 87.50 & 91.18 & 91.18 \\
$w/o$ memory       & 94.12 & \textbf{89.71} & 89.71 & 91.18 \\
\bottomrule
\end{tabular}
\end{table}

\begin{figure}[t] 
    \centering
    \begin{minipage}[b]{0.442\textwidth}
        \centering
        \setlength{\abovecaptionskip}{0pt}
        \includegraphics[width=\linewidth]{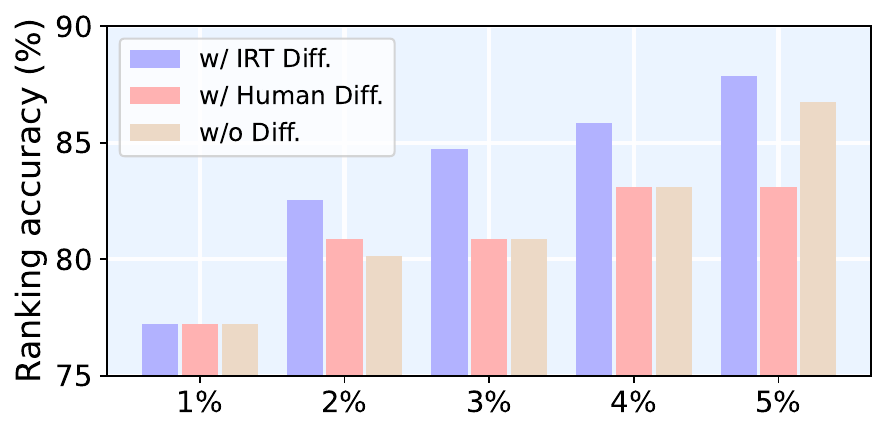}
        \caption{\textbf{Ranking accuracy of AutoJudger computed at {MMMU$_{\textit{DEV VAL}}$} under three distinct difficulty settings.}}
        \label{fig:accuracy_difficulty}
    \end{minipage}
    \hfill
    \begin{minipage}[b]{0.518\textwidth}
        \centering
        \setlength{\abovecaptionskip}{0pt}
        \includegraphics[width=\linewidth]{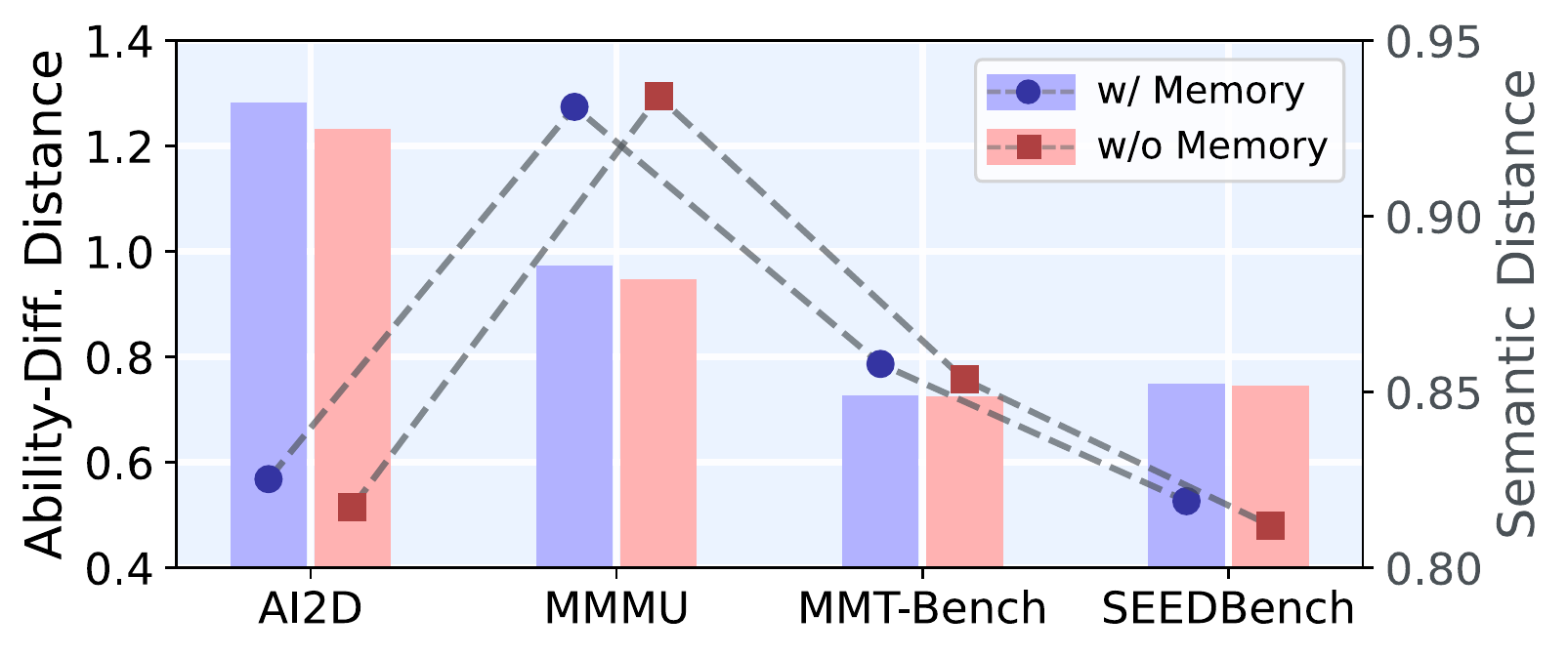}
        \caption{\textbf{Comparison of ability-difficulty distance} (bar chart, left y-axis) \textbf{and semantic distance} (line plot, right y-axis) \textbf{with and without memroy.}}
        \label{fig:dif_dist}
    \end{minipage}
\end{figure}

\subsection{Effectiveness of the AutoJudger Framework}
\label{subsection:ablation}
Moreover, we conduct experiments to validate the effectiveness of core designs within AutoJudger. 

\paragraph{Necessity of Question Difficulty Estimation.}
To investigate the necessity of providing question difficulty as reference information during the agent-driven next question selection, we conduct experiments on the MMMU dataset, which includes human-annotated difficulty labels. 

We compare three difficulty settings: (1) using IRT-estimated difficulty, (2) using human-annotated difficulty, and (3) excluding difficulty information entirely. 
Results in Figure~\ref{fig:accuracy_difficulty} illustrate that applying IRT-estimated difficulty outperforms the other two settings across various compression ratios, indicating that the difficulty information is crucial for adaptive evaluation. However, manually-assessed difficulty may not align with the difficulty perceived by existing MLLMs.

\paragraph{Necessity of Agent-based Question Selection.} 
To validate the necessity of agent-driven question selection, we conduct an ablation study by removing the agent from the AutoJudger framework. 
Candidate questions are filtered using Equation~\ref{eq:cand_subset}, and a weighted sampling mechanism is performed based on the proximity between question difficulty and the model’s estimated ability. 
As shown in the second line of Table~\ref{tab:Ablation Studies}, removing the agent leads to a significant performance drop in three datasets. The effect on SeedBench is less significant due to its large size, which causes different methods to converge. However, when the compression ratio is reduced to 1\%, the ranking accuracy drops from 86.32\% (with agent) to 84.56\% (without agent).
This demonstrates that the agent as the judger plays a crucial role in improving evaluation quality, especially in low-data scenarios.

\paragraph{Necessity of using Visual Information.}  To assess the importance of multimodal understanding in AutoJudger's decision-making process, we investigate the role of visual information in evaluation.
We construct an ablated method where visual information is excluded from the context provided to the agent, offering only textual information instead.
As presented in Table~\ref{tab:Ablation Studies}, removing visual information consistently harms the performance, indicating that the semantics of images are crucial for ensuring the diversity of selected questions in multimodal benchmarks.

\paragraph{Necessity of Dynamic Memory $\mathcal{M}$.} 
To understand the contribution of the proposed dynamic memory $\mathcal{M}_k$, we analyze the impact of removing it from the framework.
The ablated framework keeps the memory of AutoJudger empty, which transforms Equation~\eqref{eq:judgerfunc} into $q_{k+1} = f\left( \varnothing, a_{j,k}, \mathcal{C}^*_k \right)$. 

As presented in the third line in Table~\ref{tab:Ablation Studies}, the performance degrades on AI2D and MMT-Bench.
We hypothesize that without memory $\mathcal{M}$, AutoJudger relies solely on the estimated model ability $a_j$ for question selection, leading it to favor questions whose difficulty closely matches that ability.
To verify our hypothesis, we compute the averaged absolute distance $\left|d_{q_{k+1}} - a_{j,k} \right|$ between the difficulty of selected question $q_{k+1}$ and the estimated ability $a_{j,k}$ of model $m_j$. As shown in the bar chart of Figure~\ref{fig:dif_dist} (left y-axis), the ability-difficulty distance significantly decreases when memory is absent. We also compute the semantic distances between questions selected by AutoJudger. As illustrated by the line plot in Figure~\ref{fig:dif_dist} (right y-axis), semantic diversity is higher when memory is present.
These findings highlight the importance of memory in preserving a global view of model strengths and weaknesses, enabling more balanced and informative question selection.

\begin{table}[t]
\small
\centering
\caption{\textbf{Mean semantic distance between questions under 5\% compression ratio.}}
\label{mean distance}
\begin{tabular}{lcccc}
\toprule
\textbf{Method} & \textbf{AI2D}$_{\textit{TEST}}$ & {\textbf{MMMU}$_{\textit{DEV VAL}}$} & \textbf{MMT-Bench} & \textbf{SEEDBench}$_{\textit{IMG}}$ \\
\midrule
Random & 0.8169 & 0.7534 & 0.7446 & 0.7359 \\
Cluster & 0.7661 & 0.7534 & 0.7549 & 0.7402 \\
IRT & 0.8198 & 0.7568 & 0.7417 & 0.7451 \\ 
\textbf{\textit{AutoJudger}} & \textbf{0.9385} & \textbf{0.8149} & \textbf{0.8594} & \textbf{0.8262} \\
\bottomrule
\end{tabular}
\end{table}

\begin{figure}[t]
\vspace{-6pt}
  \centering
  \setlength{\abovecaptionskip}{0pt}
  \includegraphics[width=0.9\textwidth]{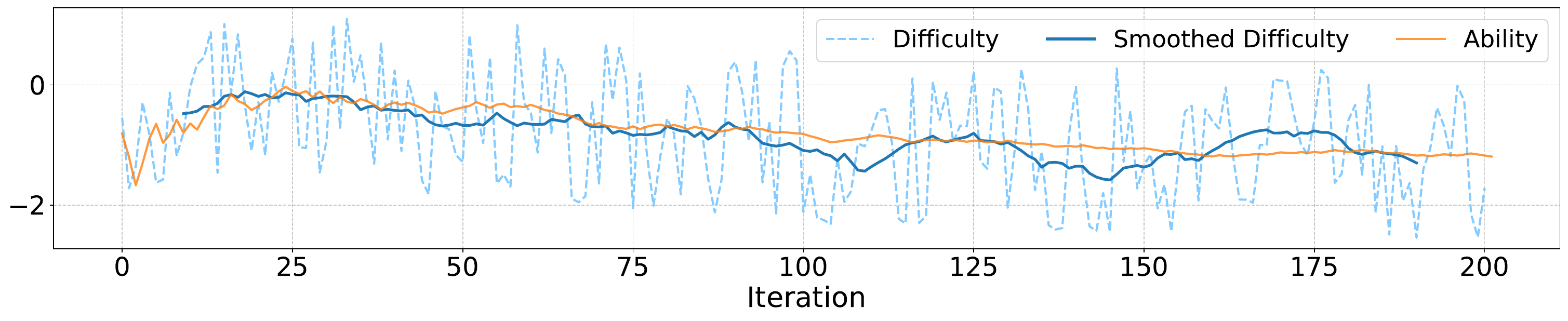}
  \caption{\textbf{The evolution of the estimated MiniCPM-V-2's ability and the question difficulty over the course of evaluation on MMMU$_{\textit{DEV VAL}}$.} ``Smoothed difficulty'' is the average difficulty of the 20 nearest questions, while ``difficulty'' is the difficulty of the question selected at each iteration.}
  \label{ability-difficulty trend}
\end{figure}

\subsection{AutoJudger Strikes a Balance between Difficulty and Semantic Diversity}
\label{subsection:further}

AutoJudger is designed to select questions that both align with the model's ability and exhibit maximal diversity. To validate this, we first investigate the relationship between the estimated model's ability and the difficulty of recommended questions. As illustrated in Figure~\ref{ability-difficulty trend}, AutoJudger adaptively selects questions whose difficulty values dynamically match the ability of the evaluated model. Beyond difficulty alignment, we then assess whether AutoJudger preserves semantic diversity in the selected questions. We quantify the semantic similarity among selected questions via the average Euclidean distance between the embeddings generated by CLIP ViT-B/32. As shown in Table~\ref{mean distance}, the questions selected by AutoJudger have a significantly higher average semantic distance than those from other methods, indicating that AutoJudger achieves superior semantic diversity.

Besides quantitative analysis, we provide several cases in Appendix~\ref{appendix:case} for qualitative analysis. These examples demonstrate that AutoJudger is capable of comprehensively analyzing both difficulty and semantics of questions, and making recommendations based on historical information.

\section{Conclusion}
\label{sec:conclusion}
We propose AutoJudger to tackle the rising cost of evaluating MLLMs. Leveraging an agent-driven question selection framework, we demonstrate that it is possible to consistently assess the capabilities of MLLMs with only 5\% samples in multimodal benchmarks. Extensive experiments illustrate that AutoJudger is not only 
effective but also stable under various evaluation settings. We believe AutoJudger is a promising solution for scalable and reliable MLLM evaluation.

\section{Broader Impacts and Limitations} 
\label{appendix:limitation}

\paragraph{Social Impacts} 
AutoJudger offers a scalable and cost-efficient framework for evaluating multimodal large language models (MLLMs), which could accelerate the development and deployment of AI systems across a wide range of applications. By drastically reducing the computational and financial overhead of benchmarking, AutoJudger lowers the barrier to entry for researchers and organizations with limited resources, promoting broader participation in AI research. Moreover, its ability to adaptively select semantically diverse and ability-matched questions may lead to more fair and comprehensive assessments of model capabilities. 

\paragraph{Limitations}
AutoJudger reduces evaluation cost by focusing on informative samples based on estimated question difficulty, avoiding the need to exhaustively test all questions. It enables efficient, targeted evaluation as models grow in scale and capability. However, this efficiency comes with a trade-off.
A primary limitation of our approach lies in the potential bias in difficulty estimation. Since question difficulty is assessed using an offline model, inaccurate estimations can affect the alignment between selected samples and the target model’s ability. 
As model capabilities continue to advance, previously challenging questions may become relatively easier, further affecting the reliability of difficulty estimates. 
To mitigate this issue, we recommend periodically updating difficulty estimates using stronger, up-to-date LLMs.


\newpage
\appendix
\section{Details of Rasch Model (IRT) Fitting}
\label{appendix:rasch_fitting}

To estimate the question difficulty vector $D$ from the binary response matrix $\{r_{ij}\}$, where $r_{ij} = 1$ indicates that model $m'_j$ correctly answers question $q_i$, we adopt the Rasch model—a one-parameter logistic IRT model as shown in Equation~\ref{eq:1PL}. 

\paragraph{Variational Bayesian Framework}
We use variational inference to approximate the posterior distribution over model abilities and question difficulties. Specifically, we assume a fully factorized variational distribution~\cite{ding2024easy2hard}:

\begin{equation}
q(a, D) = \prod_j q(a_j) \prod_i q(d_i)
\end{equation}

Each latent variable is modeled as a Gaussian:

\begin{equation}
q(a_j) = \mathcal{N}(\mu_{a_j}, \sigma^2_{a_j}), \quad q(d_i) = \mathcal{N}(\mu_{d_i}, \sigma^2_{d_i})
\end{equation}

The optimization target is the evidence lower bound (ELBO):

\begin{equation}
\mathcal{L}_\text{ELBO} = \mathbb{E}_{q(a, D)} [\log p(r \mid a, D)] - \mathrm{KL}(q(a, D) \parallel p(a, D))
\end{equation}

We adopt standard Gaussian priors: $p(a_j) = \mathcal{N}(0,1)$ and $p(d_i) = \mathcal{N}(0,10^3)$, which yield closed-form KL divergences.

\paragraph{Optimization}
We optimize the ELBO using stochastic gradient descent. Gradients are estimated via the reparameterization trick:

\begin{equation}
\begin{aligned}
a_j &= \mu_{a_j} + \sigma_{a_j} \cdot \epsilon_j,\quad \epsilon_j \sim \mathcal{N}(0, 1) \\
d_i &= \mu_{d_i} + \sigma_{d_i} \cdot \epsilon_i,\quad \epsilon_i \sim \mathcal{N}(0, 1)
\end{aligned}
\end{equation}

This leads to efficient and low-variance updates for the variational parameters $\mu$ and $\sigma$.

\paragraph{Implementation Setting}
We implement the model using PyTorch~\cite{paszke2017automatic} and Pyro\cite{bingham2017pyro}. The variational distributions over model abilities and question difficulties are initialized with zero mean and large variance. Specifically, the ability parameters $a_j$ are initialized with $\mu_{a_j} = 0$, $\sigma_{a_j} = 1$, while the difficulty parameters $d_i$ are initialized with $\mu_{d_i} = 0$, $\sigma_{d_i} = 10^3$, corresponding to vague priors that reflect minimal prior knowledge. 

We optimize the ELBO~\cite{kingma2022autoencodingvariationalbayes} using the Adam optimizer~\cite{kingma2017adammethodstochasticoptimization} with a learning rate of 0.1 for 3,200 steps, using mini-batches sampled from the response matrix $\{{r_{ij}}\}$. Training terminates when the relative change in ELBO falls below $1 \times 10^{-4}$ within a moving window. During inference, we use the variational mean $\mu_{d_i}$ as the point estimate of question difficulty.

\section{Details of Model Ability Estimation}  
\label{appendix:abi_estimation}

To estimate the model ability $a_j$  based on its responses to a subset of questions, we employ a binary search algorithm grounded in the one-parameter logistic IRT (Rasch) model. Specifically, we solve the following maximum likelihood estimation problem:
\begin{equation}
    \max_{a_j} \sum_{i \in Q'} \log p(r_{ij} \mid a_j, d_i),
\end{equation}
where $p(r_{ij} \mid a_j, d_i)$ is defined in Equation~\ref{eq:1PL}. Since the log likelihood is a monotonic function with respect to $a_j$, we perform binary search within a bounded interval $[-30, 30]$, iteratively updating the estimate until convergence. The stopping criterion is based on a fixed threshold of $10^{-5}$  for either the log-likelihood difference or the change in $a_j$.

This procedure enables efficient and stable estimation of real-time model ability during evaluation, while keeping the question difficulties $\{d_i\}$ fixed.

\section{Implementation Details of Efficient Benchmarking Methods}
\label{appendix:implementation}

\subsection{Baselines}
We detail the implementation of the baselines below.
\begin{itemize}
    \item \textit{Random Sampling (Random)}: We uniformly sample $\delta*|Q|$ questions from the complete evaluation benchmark $Q$ without replacement.
    \item \textit{Stratified Random Sampling (Stratified)}~\cite{perlitz2023efficient}: We partition the question pool based on provided category labels and draw approximately equal numbers of questions from each partition. We ensure that the maximum size difference between any two categories is no greater than one. Sampling is performed independently per category without replacement.
    \item \textit{Cluster-Based Sampling (Cluster)}: Each question is embedded using the CLIP ViT-B/32 encoder~\cite{radford2021learning}, producing a 512-dimensional representation. Embeddings are L2-normalized before clustering. We apply K-means clustering to partition the question pool. The number of clusters $K$ is set to the number of desired questions, i.e., $K = \delta \cdot |Q|$. One question is selected per cluster, chosen as the one closest to the centroid in Euclidean space.
    \item \textit{Optimal IRT Difficulty Choosing (IRT)}~\cite{lord2012applications}: We use a one-parameter logistic Item Response Theory model to adaptively select questions based on the model's estimated ability. The ability score is initialized with a simple prior: we assume the model has answered five medium-difficulty questions (difficulty 0) and got 2.5 correct on average. This initialization prevents unstable updates in early iterations.
\end{itemize}

\subsection{Our Framework: AutoJudger}

The evaluation workflow of AutoJudger is presented in Algorithm~\ref{alg:autojudger}. All questions in each benchmark are first annotated with estimated difficulty levels. Then, evaluation begins with a standardized initialization, followed by iterative refinement of the question set based on the model’s responses.

\paragraph{Training and Test Models} 
To ensure a representative and balanced evaluation, we partition the models based on parameter scale, as the model capability is generally observed to improve with increasing parameter size. Accordingly, we divide the models into four groups: $<5B$, $<9B$, $<16B$, and $\geq16B$ (including proprietary models). From each group, we randomly sample 20\% of the models as test models, the remaining 80\% are used as training models to collect offline responses which are utilized for question difficulty estimation. This stratified selection strategy ensures that AutoJudger is evaluated across a wide spectrum in terms of model abilities. The complete list of the 60 training models used for IRT-based question difficulty assessment are provided in Table~\ref{tables:trainset} and the left 17 models evaluated via AutoJudger are listed in Table~\ref{tables:testset}. 

\begin{algorithm}
\caption{\textbf{AutoJudger} - Adaptive Ability Estimation via Multimodal Large Language Model}
\label{alg:autojudger}
\begin{algorithmic}[1]
\Require 
    \State Question Pool $Q=\{q_i\}$, Testing Models $M=\{m_j\}$
    \State An $Agent$ as the judger to select the question

\ForAll{test model $m_j \in M$}
    \State Initialize problem assistant $Agent$
    \State Sample initial batch of questions $Q_0$
    \State collect model responses $R_{0}$
    \State Estimate initial model ability $a_0$.
    \State Generate init memory $\mathcal{M}_0$.

    \For{$t = 1$ to $(\delta\times|Q|-|Q_0|)$}
        \State Get the candidate set $\mathcal{C}^*$
        \State $Agent$ select the next question with $\mathcal{M}_{t-1},a_{t-1}$ given. ($q_t=f(\mathcal{M}_{t-1},a_{t-1},\mathcal{C}^*)$)
        \State Get the answer $r_{tj}$, update the ability $a_t$
        \State Update $\mathcal{M}_t$ with $q_t$ included.
    \EndFor
    \State Get the final ability estimation $a_{(\delta\times|Q|-|Q_0|)}$
\EndFor

\end{algorithmic}
\end{algorithm}

\newpage

\begin{table}[h]
\centering
\small
\caption{\textbf{List of 60 training set models used for IRT-based question difficulty assessment.} These models span a range of sizes and include both open-source and proprietary models.}
\label{tables:trainset}
\resizebox{0.67\linewidth}{!}{
\begin{tabular}{cccc}
\toprule
Models & Open-source & \# Params (B) & Date \\ \midrule
InternVL2-1B~\cite{chen2024internvl} & Yes & 0.9 & 2024.11 \\
llava-onevision-qwen2-0.5B-ov~\cite{li2024llavaonevisioneasyvisualtask} & Yes & 0.9 & 2024.07 \\
llava-onevision-qwen2-0.5B-si~\cite{li2024llavaonevisioneasyvisualtask} & Yes & 0.9 & 2024.07 \\
h2ovl-mississippi-1B~\cite{galib2024h2ovl} & Yes & 0.8 & 2024.01 \\
NVLM~\cite{dai2024nvlm} & Yes & 79.4 & 2024.09 \\
Qwen2-VL-72B-Instruct~\cite{wang2024qwen2} & Yes & 72 & 2024.08 \\
360VL-70B~\cite{360vl_70b_2024} & Yes & 71 & 2024.04 \\
InternVL2-40B~\cite{chen2024internvl} & Yes & 40.1 & 2024.06 \\
InternVL-Chat-V1-5~\cite{chen2024internvl} & Yes & 25.5 & 2024.03 \\
InternVL2-26B~\cite{chen2024internvl} & Yes & 25.5 & 2024.11 \\
MMAlaya2~\cite{datacanvas2024mmalaya2} & Yes & 25.5 & 2024.08 \\
Eagle-X5-13B~\cite{shi2024eagle} & Yes & 15.4 & 2024.08 \\
Slime-13B~\cite{zhang2024beyond} & Yes & 13.4 & 2024.05 \\
TransCore-M~\cite{transcorem_2023} & Yes & 13.4 & 2024.03 \\
llava-v1.5-13B~\cite{liu2024improvedbaselinesvisualinstruction} & Yes & 13 & 2024.01 \\
Falcon2-VLM-11B~\cite{falcon_11b_vlm_2024} & Yes & 11 & 2024.07 \\
Ovis1.6-Gemma2-9B~\cite{lu2024ovisstructuralembeddingalignment} & Yes & 10.2 & 2024.09 \\
monkey~\cite{li2024monkeyimageresolutiontext} & Yes & 9.8 & 2023.11 \\
monkey-chat~\cite{li2024monkeyimageresolutiontext} & Yes & 9.8 & 2023.11 \\
POINTS-Yi-1.5-9B-Chat~\cite{liu2024pointsimprovingvisionlanguagemodel} & Yes & 9.5 & 2024.09 \\
Mantis-8B-Fuyu~\cite{Jiang2024MANTISIM} & Yes & 9.4 & 2024.04 \\
Eagle-X5-7B~\cite{shi2024eagle} & Yes & 9.1 & 2024.08 \\
Bunny-llama3-8B~\cite{he2024efficientmultimodallearningdatacentric} & Yes & 8.5 & 2024.04 \\
Mantis-8B-siglip-llama3~\cite{jiang2024mantisinterleavedmultiimageinstruction} & Yes & 8.5 & 2024.04 \\
Mantis-8B-Idefics2~\cite{jiang2024mantisinterleavedmultiimageinstruction} & Yes & 8.4 & 2024.05 \\
Slime-8B~\cite{zhang2024llavahddivinghighresolutionlarge} & Yes & 8.4 & 2024.05 \\
llava-next-llama3~\cite{liu2024llavanext} & Yes & 8.3 & 2024.04 \\
POINTS-Qwen-2.5-7B-Chat~\cite{liu2024pointsimprovingvisionlanguagemodel} & Yes & 8.3 & 2024.12 \\
llava-next-interleave-7B~\cite{liu2024llavanext} & Yes & 8.1 & 2024.06 \\
llava-next-interleave-7B-dpo~\cite{liu2024llavanext} & Yes & 8.1 & 2024.06 \\
MiniCPM-V-2-6~\cite{yao2024minicpm} & Yes & 8.1 & 2024.07 \\
InternVL2-8B~\cite{chen2024far} & Yes & 8.1 & 2024.11 \\
llava-onevision-qwen2-7B-ov~\cite{li2024llavaonevisioneasyvisualtask} & Yes & 8.0 & 2024.07 \\
Ovis1.5-Llama3-8B~\cite{lu2024ovisstructuralembeddingalignment} & Yes & 8 & 2024.07 \\
molmo-7B-O-0924~\cite{deitke2024molmopixmoopenweights} & Yes & 7.7 & 2024.09 \\
llava-next-mistral-7B~\cite{liu2024llavanext} & Yes & 7.6 & 2024.03 \\
deepseek-vl-7B~\cite{lu2024deepseekvl} & Yes & 7.3 & 2024.02 \\
llava-next-vicuna-7B~\cite{liu2024llavanext} & Yes & 7.1 & 2024.05 \\
XComposer2~\cite{dong2024internlmxcomposer2masteringfreeformtextimage} & Yes & 7 & 2024.01 \\
llava-v1.5-7B~\cite{liu2024improvedbaselinesvisualinstruction} & Yes & 7 & 2024.01 \\
Phi-3-Vision~\cite{abdin2024phi} & Yes & 4.2 & 2024.05 \\
InternVL2-4B~\cite{chen2024far} & Yes & 3.7 & 2024.11 \\
Vintern-3B-beta~\cite{doan2024vintern1befficientmultimodallarge} & Yes & 3.2 & 2024.01 \\
BlueLM-V~\cite{lu2024bluelmv3balgorithmcodesignmultimodal} & No & 3 & 2024.11 \\
paligemma-3B-mix-448~\cite{beyer2024paligemma} & Yes & 2.9 & 2024.04 \\
InternVL2-2B~\cite{chen2024far} & Yes & 2.2 & 2024.11 \\
Aquila-VL-2B~\cite{gu2024infinitymmscalingmultimodalperformance} & Yes & 2.2 & 2024.01 \\
deepseek-vl-1.3B~\cite{lu2024deepseekvlrealworldvisionlanguageunderstanding} & Yes & 2.0 & 2024.02 \\
Moondream1~\cite{MoonDream} & Yes & 1.9 & 2024.01 \\
XComposer2-1.8B~\cite{dong2024internlmxcomposer2masteringfreeformtextimage} & Yes & 1.8 & 2024.01 \\
Kosmos2~\cite{peng2023kosmos} & Yes & 1.7 & 2023.06 \\
molmoE-1B-0924~\cite{deitke2024molmo} & Yes & 1 & 2024.09 \\
GPT4V-20240409-HIGH~\cite{gpt-4v} & No & - & 2024.04 \\
GPT4o~\cite{hurst2024gpt} & No & - & 2024.05 \\
GPT4o-HIGH~\cite{hurst2024gpt} & No & - & 2024.05 \\
GeminiFlash1-5~\cite{team2024gemini} & No & - & 2024.09 \\
JT-VL-Chat~\cite{JTVL} & No & - & 2024.10 \\
Qwen-VL-Max-0809~\cite{bai2023qwenvlversatilevisionlanguagemodel} & No & - & 2024.08 \\
Qwen-VL-Plus-0809~\cite{bai2023qwenvlversatilevisionlanguagemodel} & No & - & 2024.08 \\
Taiyi~\cite{luo2024taiyi} & No & - & 2023.11 \\
\bottomrule
\end{tabular}
}
\end{table}

\newpage

\begin{table}[t]
\small
\centering
\caption{\textbf{List of 17 test set models evaluated using AutoJudger.} These models are disjoint from the training set and representative in terms of diverse types and scales.}
\label{tables:testset}
\begin{tabular}{cccc}
\toprule
Models & Open-source & \# Params (B) & Date \\ \midrule
InternVL2-76B~\cite{chen2024expanding} & Yes & 76.3 & 2024.06 \\
llava-next-vicuna-13B~\cite{liu2024llavanext} & Yes & 13.4 & 2024.02 \\
Pixtral-12B~\cite{agrawal2024pixtral} & Yes & 12 & 2024.08 \\
Ovis1.5-Gemma2-9B~\cite{lu2024ovisstructuralembeddingalignment} & Yes & 11.4 & 2024.07 \\
idefics2-8B~\cite{laurenccon2024matters} & Yes & 8.4 & 2024.03 \\
Mantis-8B-clip-llama3~\cite{Jiang2024MANTISIM} & Yes & 8.3 & 2024.01 \\
llava-onevision-qwen2-7B\_si~\cite{li2024llavaonevisioneasyvisualtask} & Yes & 8.0 & 2024.07 \\
molmo-7B-D-0924~\cite{deitke2024molmopixmoopenweights} & Yes & 8.0 & 2024.09 \\
Slime-7B~\cite{zhang2024beyond} & Yes & 7.1 & 2024.05 \\
Ovis1.6-Llama3.2-3B~\cite{lu2024ovisstructuralembeddingalignment} & Yes & 4.1 & 2024.01 \\
MiniCPM-V-2~\cite{yao2024minicpm} & Yes & 3.4 & 2024.11 \\
h2ovl-mississippi-2B~\cite{galib2024h2ovl} & Yes & 2.1 & 2024.01 \\
Janus-1.3B~\cite{wu2024janus} & Yes & 2.1 & 2024.01 \\
Moondream2~\cite{moondream2} & Yes & 1.9 & 2024.02 \\
GPT4o-20240806~\cite{hurst2024gpt} & No & - & 2024.08 \\
GeminiPro1-5~\cite{team2024gemini} & No & - & 2024.09 \\
Step1V~\cite{step-1v} & No & - & 2024.03 \\
\bottomrule
\end{tabular}
\end{table}

\paragraph{Initialization Details} 

In the initialization phase, we take the text of each question as input and use CLIP ViT-B/32 as the encoding model (this design choice is provided in Appendix~\ref{appendix:initialization}) to generate normalized vector representations. We aim to select a diverse and representative set of questions, so we apply k-means clustering with k=10 and select the questions closest (in terms of L2 distance) to each of the resulting cluster centers. To mitigate the instability of k-means, we repeat the clustering process 50 times and choose the set that achieves the highest ranking accuracy on the training set.

\section{Further Analysis}

\begin{table}[t]
\centering
\caption{\textbf{Comparison between different initialization methods.} ``Multi-concat'' and ``multi-mean'' refer to concatenating and averaging the image and text embeddings, respectively, to serve as the multi-modal representations of questions. We report the average rank of each method (among all methods) across four benchmarks as the overall performance.}
\label{tab:init_compare}
\resizebox{\linewidth}{!}{
\begin{tabular}{llccccc}
\toprule
{Encoding Models} & {Input} 
& {$\text{AI2D}_{\textit{TEST}}$} & MMMU$_{\textit{DEV VAL}}$ & {$\text{MMT-Bench}$} & SEEDBench$_{\textit{IMG}}$ & {Avg Rank} \\
\midrule

\multirow{4}{*}{CLIP\,ViT-B/32} 
& multi-concat & 74.98 & 72.54 & 61.81 & 58.00 & 7.75 \\
& multi-mean   & 72.61 & 70.57 & 55.07 & 55.92 & 10.00 \\
& image       & 69.80 & 64.72 & 66.07 & 65.45 & 7.50 \\
& text        & 81.31 & 73.02 & 60.88 & 75.59 & 3.75 \\
\midrule

\multirow{4}{*}{CLIP\,ViT-L/14}
& multi-concat & 67.23 & 75.53 & 52.00 & 61.61 & 9.50 \\
& multi-mean   & 75.70 & 75.79 & 67.00 & 54.88 & 5.25 \\
& image       & 73.24 & 74.52 & 62.30 & 64.01 & 6.00 \\
& text        & 77.45 & 66.38 & 66.15 & 62.62 & 6.25 \\
\midrule

\multirow{3}{*}{Qwen2.5-VL-7B}
& text\&image      & 78.12 & 62.77 & 62.02 & 67.91 & 6.25 \\
& image         & 70.77 & 64.77 & 63.55 & 64.53 & 8.00 \\
& text           & 79.58 & 66.15 & 59.82 & 64.53 & 7.25 \\
\midrule
\multirow{1}{*}{IRT}
& -           & 80.40 & 70.40 & 61.81 & 64.75 & 5.00 \\
\multirow{1}{*}{Random}
& -           & 74.62 & 68.83 & 60.36 & 63.33 & 8.50 \\
\bottomrule

\end{tabular}}
\end{table}

\subsection{The Impact of Initialization}
\label{appendix:initialization}

Initialization is a crucial component of the AutoJudger framework, as it determines the starting point for iterative evaluation. To systematically investigate its impact, we conduct experiments to compare the performance of different initialization methods, including two baselines: random sampling and sampling based on question difficulty quantiles (denoted as IRT), as well as our clustering-based initialization. Additionally, we compare the performance of different embedding strategies, including the use of different encoders and different semantic representations of questions.

To reduce the impact of randomness, each configuration is evaluated over 50 independent trials, and the average ranking accuracy is reported. To avoid information leakage and overfitting to the test set, we report the ranking accuracy on the training-set models.
As shown in Table~\ref{tab:init_compare}, there are several findings: 
(1) Incorporating additional information—whether related to difficulty or semantics—through appropriate methods improves the effectiveness of initialization.
(2) While approaches based on difficulty perform well, using semantic information yields the best results.
(3) Unlike Qwen2.5-VL, the dual-encoder CLIP model struggles to integrate information from multiple modalities.
(4) Textual features reflect the diversity of questions more effectively than visual features, suggesting that current benchmarks may not consider the richness of visual information.

\subsection{The Impact of Candidate Question Retrieval Strategy} 
\begin{table}[t]
\small
\centering
\caption{\textbf{Ranking accuracy for top-performing models with personalized and unified question selection strategies.} ``personalized'' means the questions are selected via Equation~\ref{eq:cand_subset}. ``simplest'' means the questions are from the evaluation of the lowest-ranked model and fixed for all models.}

\label{tab:appD2 personalized-unify}
\begin{tabular}{lc|ccccc}
\toprule
\multirow{2}{*}{Benchmark} & \multirow{2}{*}{Questions} & \multicolumn{5}{c}{Compression Ratio} \\ \cmidrule{3-7}
& & 1\% & 2\% & 3\% & 4\% & 5\% \\ 
\midrule
\multirow{2}{*}{$\text{AI2D}_{\textit{TEST}}$} 
& personalized & \textbf{95.24} & \textbf{95.24} & {95.24} & \textbf{95.24} & \textbf{95.24} \\
& simplest & 85.71 & 90.48 & \textbf{100.0} & \textbf{95.24} & 90.48 \\
\midrule
\multirow{2}{*}{MMMU$_{\textit{DEV VAL}}$}

& personalized & \textbf{78.57} & \textbf{81.25} & \textbf{82.14} & \textbf{88.39} & \textbf{88.39} \\
& simplest & \textbf{78.57} & 80.00 & 80.71 & 85.71 & 85.71 \\

\midrule

\multirow{2}{*}{$\text{MMT-Bench}$}

& personalized & \textbf{86.61} & \textbf{86.61} & \textbf{90.18} & \textbf{92.86} & \textbf{93.75} \\
& simplest & 85.71 & 82.14 & 78.57 & 65.71 & 67.14 \\

\bottomrule
\end{tabular}
\end{table}
\paragraph{Superiority of Personalized Retrieval} 
As argued in the introduction, we believe each model should be assigned with a personalized evaluation subset since models vary in capability. For instance, evaluating powerful models with too many easy questions may provide limited information.
To validate this argument, we conduct an experiment to assess top-performing models (top 50\% in terms of average ranks), either with their personalized questions picked by AutoJudger or simple questions that are selected to evaluate the worst model.

Results are provided in Table~\ref{tab:appD2 personalized-unify}. Considering the efficiency, SEEDBench is excluded in this experiment due to its large scale. 
Since AI2D is a relatively easy scenario (minimal variation in difficulty across the questions), the ``simplest'' strategy demonstrates comparative performance. However, on more complex benchmarks like MMT and MMMU, the personalized approach demonstrates superior performance. Additionally, we observed that the ``simplest'' strategy lacks stability and does not necessarily improve as the dataset size increases.
Generally, by dynamically selecting questions tailored to each model’s capability, the proposed AutoJudger framework better accommodates varying model strengths and avoids overfitting to the preference of specific models. Therefore, we adopt the personalized strategy to retrieve questions.

\begin{table}[t]
\small
\centering
\caption{\textbf{Comparison of different candidate question selection strategies. }"Semantic farthest" means selecting questions with the largest semantic distance, "optimal difficulty" means selecting questions with the smallest difficulty distance, and "random" means purely random selection.}

\label{tab:app_candidate_strategy}
\begin{tabular}{lc|ccccc}
\toprule
\multirow{2}{*}{Benchmark} & \multirow{2}{*}{Strategy} & \multicolumn{5}{c}{Compression Ratio} \\ \cmidrule{3-7}
& & 1\% & 2\% & 3\% & 4\% & 5\% \\ 
\midrule
\multirow{3}{*}{{$\text{AI2D}_{\textit{TEST}}$}} 
& semantic farthest & 85.29 & 89.71 & 93.38 & 92.65 & 94.85 \\
& optimal difficulty & \textbf{90.44} & \textbf{93.38} & \textbf{93.38} & \textbf{94.12} & \textbf{95.59} \\
& random & 86.76 & 91.18 & 91.18 & 91.91 & 91.18 \\
\midrule
\multirow{3}{*}{MMMU$_{\textit{DEV VAL}}$}
& semantic farthest & 77.21 & 82.54 & 84.74 & \textbf{85.85} & \textbf{87.94} \\
& optimal difficulty & 77.21 & 77.21 & 78.68 & 80.15 & 83.09 \\
& random & 77.21 & \textbf{83.09} & \textbf{86.03} & {83.82} & {83.82} \\
\midrule
\multirow{3}{*}{$\text{MMT-Bench}$}
& semantic farthest & \textbf{85.66} & \textbf{87.87} & \textbf{89.34} & \textbf{91.91} & \textbf{92.06} \\
& optimal difficulty & 72.06 & 78.68 & 78.68 & 83.09 & 82.35 \\
& random & {77.94} & {83.82} & 88.97 & 86.76 & 91.18 \\
\midrule
\multirow{3}{*}{SEEDBench$_{\textit{IMG}}$}
& semantic farthest & \textbf{86.40} & \textbf{88.24} & \textbf{89.89} & \textbf{90.99} & \textbf{90.74} \\
& optimal difficulty & 72.06 & 78.68 & 78.68 & 83.09 & 82.35 \\
& random & 78.68 & 83.82 & {81.68} & {83.82} & {88.97} \\
\bottomrule
\end{tabular}
\end{table}

\paragraph{Candidate Question Selection Strategy}
As stated in Equation~\ref{eq:cand_num} in Section~\ref{subsection:retrieval}, we select questions with largest semantic distance as the candidates (``semantic farthest''). To demonstrate the superiority of our approach, we compare it against two widely adopted question selection baselines:  random sampling (``random''), and selecting questions with the smallest difficulty distance (``optimal difficulty''). As summarized in Table~\ref{tab:app_candidate_strategy}, while the ``optimal difficulty'' strategy achieves the best performance on the AI2D$_{\text{TEST}}$ benchmark, its effectiveness does not generalize well across other benchmarks. In contrast, the ``semantic farthest'' strategy demonstrates consistently strong performance across all evaluated benchmarks and under different compression ratios. Therefore, we choose to use semantic farthest strategy, as it not only exhibits broad applicability and consistent performance across diverse benchmarks, but can also introduce greater informational diversity.

\begin{table}[t]
\small
\centering
\caption{\textbf{The impact of different number of candidate questions.}}
\label{tab:appD2}

\begin{tabular}{lc|ccccc}
\toprule
\multirow{2}{*}{Benchmark} & \multirow{2}{*}{\# Candidate} & \multicolumn{5}{c}{Compression Ratio} \\ \cmidrule{3-7}
& & 1\% & 2\% & 3\% & 4\% & 5\% \\
\midrule
\multirow{3}{*}{{$\text{AI2D}_{\textit{TEST}}$}} 

& 5 & 85.29 & 89.71 & \textbf{93.38} & 92.65 & \textbf{94.85} \\
& 7 & \textbf{88.24} & \textbf{92.65} & \textbf{93.38} & \textbf{93.38} & \textbf{94.85} \\
& 10 & 83.82 & 86.03 & 89.71 & 91.18 & 91.91 \\
\midrule

\multirow{3}{*}{MMMU$_{\textit{DEV VAL}}$}

& 5 & \textbf{77.21} & \textbf{82.54} & \textbf{84.74} & \textbf{85.85} & \textbf{87.94} \\
& 7 & \textbf{77.21} & 82.35 & 80.88 & 83.09 & 83.82 \\
& 10 & \textbf{77.21} & {82.35} & {83.82} & {85.29} & {85.29} \\
\midrule

\multirow{3}{*}{$\text{MMT-Bench}$}

& 5 & \textbf{85.66} & 87.87 & \textbf{89.34} & \textbf{91.91} & 92.06 \\
& 7 & 83.09 & \textbf{88.24} & 88.24 & 88.97 & \textbf{92.65} \\
& 10 & 82.35 & 86.03 & 88.24 & {91.18} & {91.18} \\
\midrule

\multirow{3}{*}{SEEDBench$_{\textit{IMG}}$}

& 5 & \textbf{86.40} & \textbf{88.24} & \textbf{89.89} & {90.99} & 90.74 \\
& 7 & 84.56 & 86.76 & 88.97 & \textbf{92.65} & \textbf{92.65} \\
& 10 & 86.03 & 86.76 & 88.24 & 88.24 & 88.97 \\
\bottomrule
\end{tabular}
\end{table}

\paragraph{Expansion of Candidate Question Pool} 
We investigate the impact of the number of candidate questions (see Equation~\ref{eq:cand_num}) on the performance of AutoJudger by expanding $|C_k^*|$ from 5 to 7 and 10.
As shown in Table~\ref{tab:appD2}, a larger candidate set introduces more flexibility, but also brings additional noise, making it harder to identify the optimal next question. 
It also increases the context length, posing challenges for AutoJudger. After comparing across different configurations, we find that using 5 candidates offers a good trade-off and set $|C_k^*| = 5$ as the default in our experiments.

\begin{figure}[t]
    \centering
    \includegraphics[width=0.9\textwidth]{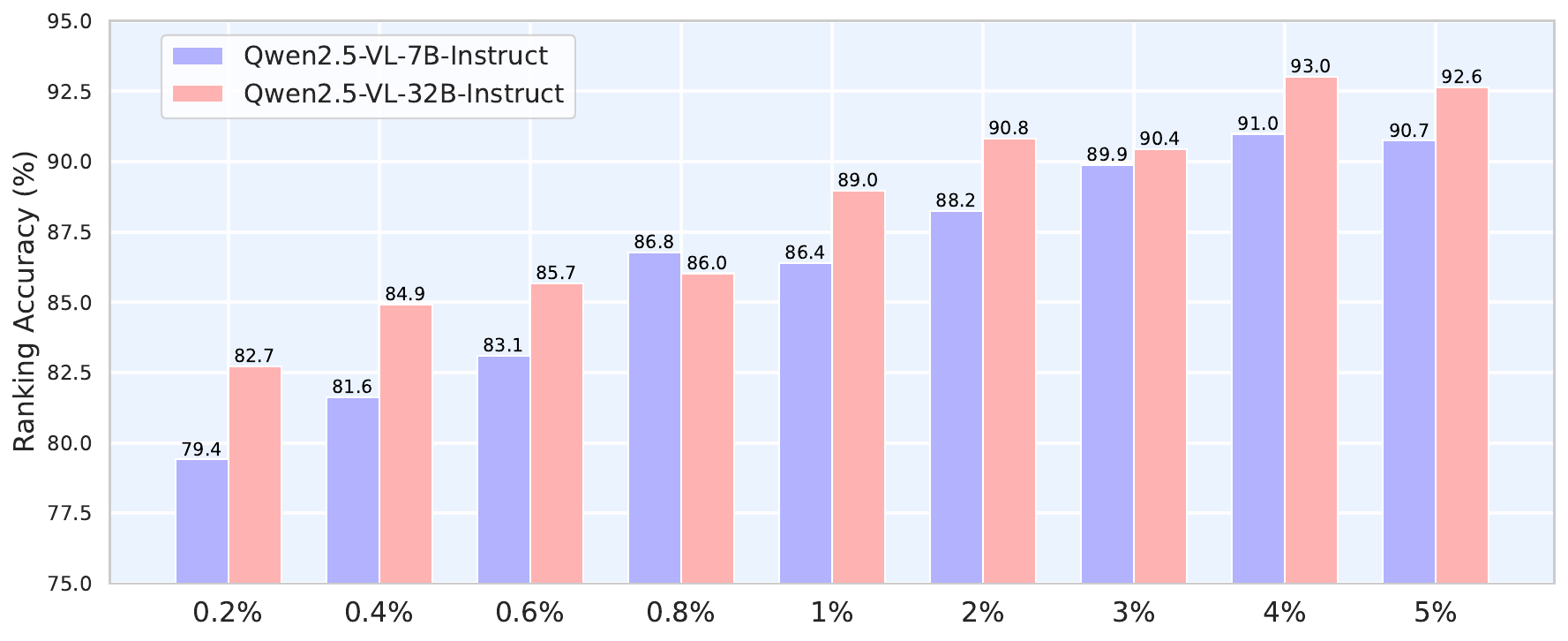}  
    \caption{Comparison of 7B and 32B models on SEEDBench$_{\textit{IMG}}$ at different compression ratios.}
    \label{fig:Choice Agent}
\end{figure}

\subsection{Scaling up the Judging Agent} 
As the capability of the interviewer/judging agent plays a crucial role in our framework. We explore to replace the original 7B-scale agent with a larger backbone, i.e. Qwen2.5-VL-32B-Instruct, to assess the impact of the scale of judging agent. 
Figure~\ref{fig:Choice Agent} presents a comparison between the 7B and 32B models on SEEDBench under varying compression ratios. The results show that the 32B model consistently outperforms the 7B models, especially in low-ratio settings. 
For example, at a compression ratio of 0.2\%, the 32B model achieves a ranking accuracy of 82.7\%, representing a 3.3\% improvement over the 7B model.
Although both models improve as the compression ratio increases, the 32B model remains consistently stronger, demonstrating better generalization and selection capability. 
The above findings indicate that the AutoJudger framework not only demonstrates strong performance, but also shows potential for further expansion. A more advanced interviewer agent could further enhance the effectiveness of AutoJudge.

\section{Case Study}
\label{appendix:case}

\begin{figure}[h]
  \centering
  \setlength{\abovecaptionskip}{0pt}
  \includegraphics[width=1.0\linewidth]{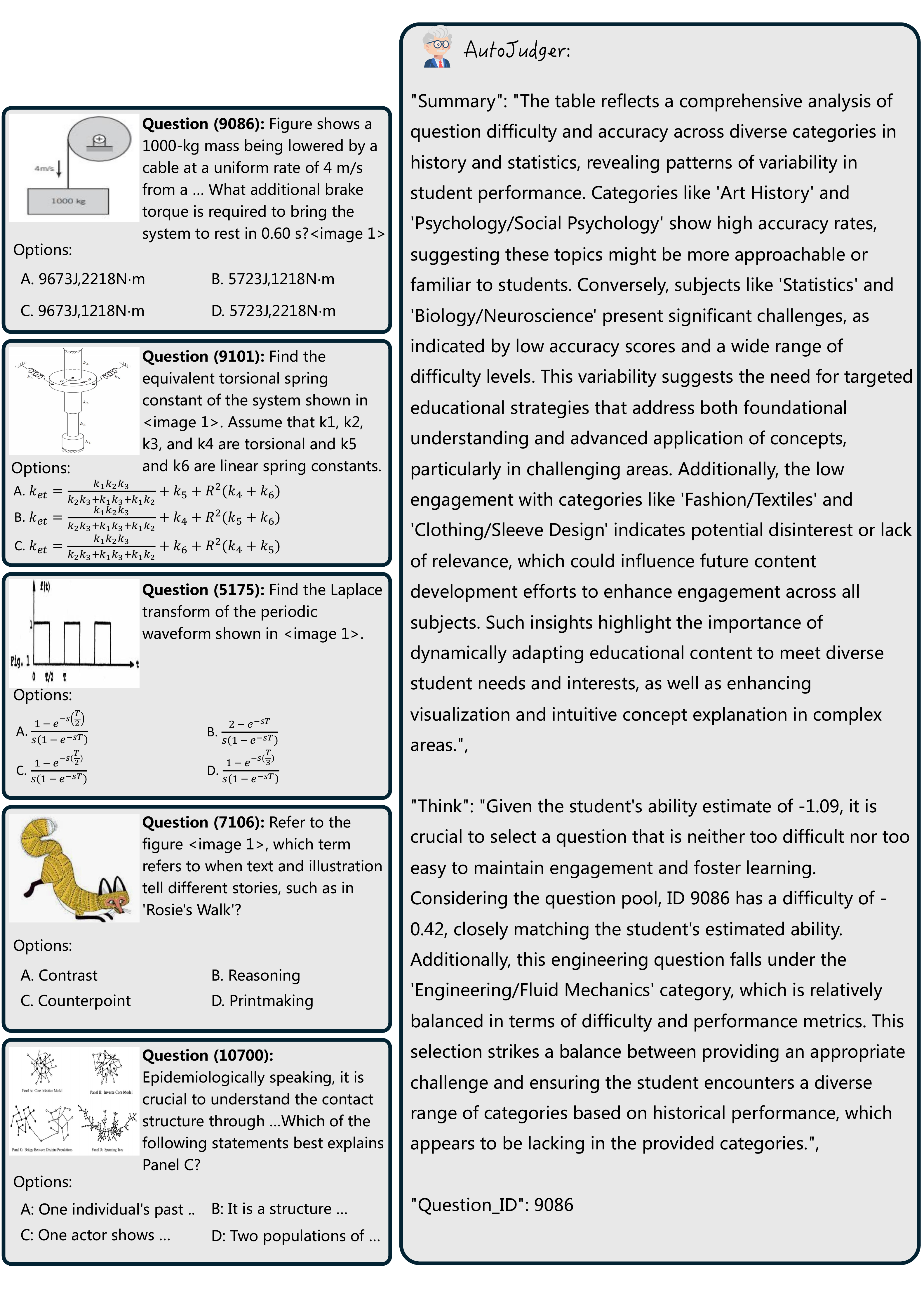}
  \caption{Response Examples from AutoJudger on MMMU.}
  \label{case_MMMU}
\end{figure}

\begin{figure}[h]
  \centering
  \setlength{\abovecaptionskip}{0pt}
  \includegraphics[width=1.0\linewidth]{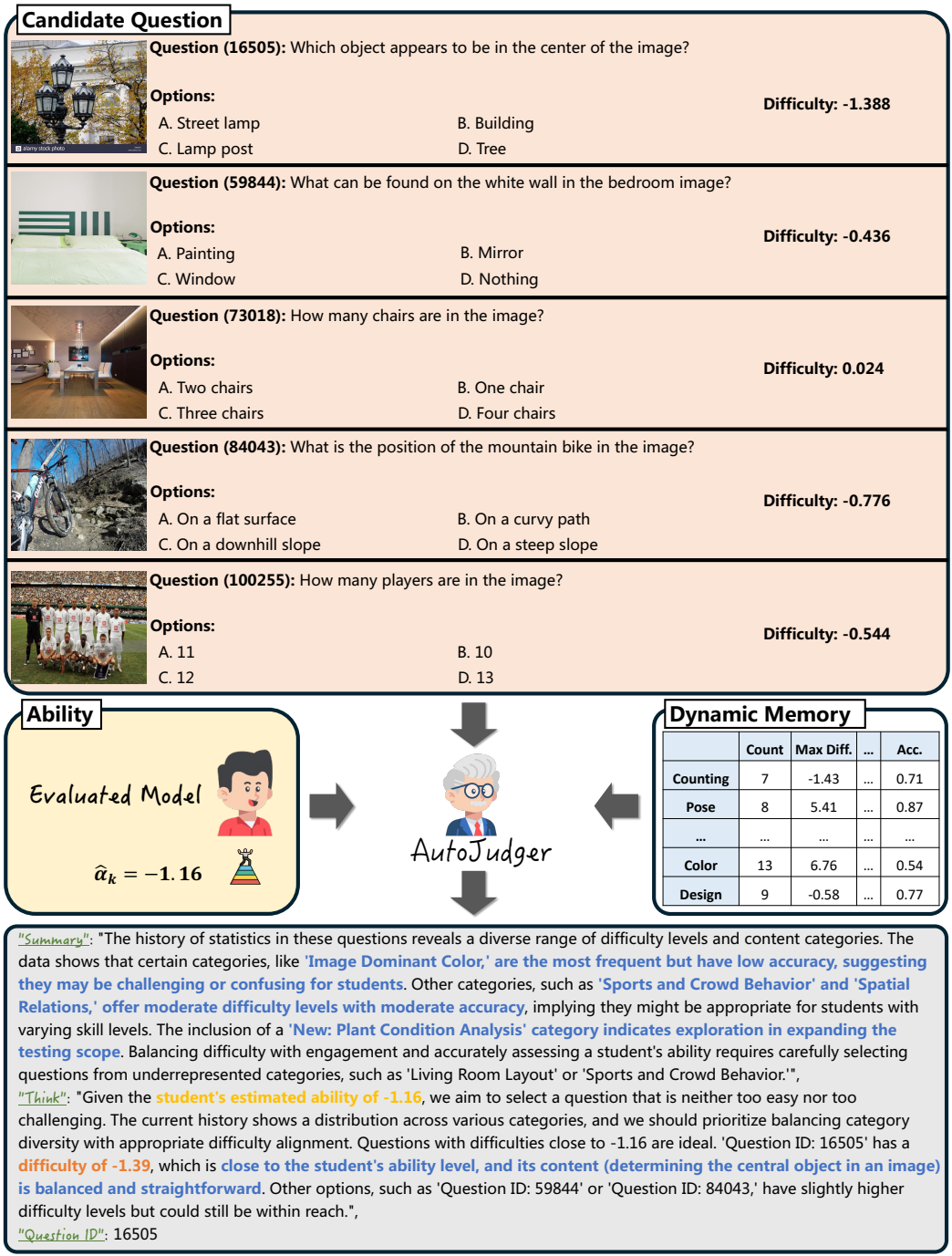}
  \caption{Response Examples from AutoJudger on SeedBench.}
  \label{case_SEEDBENCH}
\end{figure}

As an agent-driven evaluation framework, AutoJudger offers a major advantage in enhancing the interpretability of assessment results. We provide two representative examples in Figure~\ref{case_MMMU} and Figure~\ref{case_SEEDBENCH}. These cases illustrate how information stored in the dynamic memory enables the agent to efficiently analyze the evaluated model’s performance across different types of questions (highlighted in blue text in the figures), thereby guiding more informed selection of subsequent evaluation items. Furthermore, the combination of model ability estimation and corresponding question difficulty analysis (marked in yellow and orange) assists the agent in identifying the most appropriate questions. Supported by these key components, AutoJudger can not only evaluate models efficiently, but also provide transparent reasoning behind each evaluation decision. We believe this is an essential step toward building trustworthy and transparent evaluation frameworks for future AI systems.

\section{Prompt of AutoJugder}
\label{appendix:prompt}
\renewcommand{\lstlistingname}{Prompt}
\begin{tcolorbox}[width=\linewidth]
\begin{lstlisting}[caption=Category Identification for Initialization Stage ]

You are an expert educational AI assistant specializing in question classification. 
Your task is to analyze the provided questions and categorize them into meaningful subject/topic categories.

Task Overview:
You will analyze a set of practice questions (including both text and images) and classify each question into a meaningful category(Expect two or more question in the same category).
The output should be a JSON object mapping question IDs to their respective categories.
{
Question ID: # Question ID
Difficulty: # Difficulty
Content: # Content
# IAMAG
Options: # Options
...
}

Output Requirements:
- Return a JSON object with the following format:
{
  "<Question_ID_1>": "<Category_Name_1>",
  "<Question_ID_2>": "<Category_Name_2>",
  ...
}
- Keys are question IDs (index) from the input data.
- Values are descriptive category names that you assign.
- ONLY return the JSON object; do not include any other text or explanation.

\end{lstlisting}
\end{tcolorbox}

\begin{tcolorbox}[width=\linewidth]
\begin{lstlisting}[caption=Category Identification for Iteration Stage ]
You are an expert educational classifier. Analyze the question and determine its category.
{
Question ID: # Question ID
Difficulty: # Difficulty
Content: # Content
# IAMAG
Options: # Options
...
}

Task: Review the question above. Determine all applicable categories from the existing list: {# Category}, or include new categories if necessary.

Output Requirements:
- Return a JSON object with:
{"category": ["Existing or new category name(s)"]}
- List ALL relevant categories (minimum 1 item).
- Use EXACT names for existing categories.
- Include multiple entries if needed (e.g., mixed existing/new categories).
- Do NOT add explanations, only JSON.

\end{lstlisting}
\end{tcolorbox}

\begin{tcolorbox}[width=\linewidth]
\begin{lstlisting}[caption=Question Recommendend]

You are an expert educational AI assistant. 
Your task is to select the most appropriate next question from the candidate pool based on:
1. The student's current ability (# ability) estimated by IRT.
2. The diversity of question categories in the history.
3. The match between question difficulty and student ability.
Prioritize questions that balance category diversity and difficulty alignment.

Statistics in history questions
{
# Memory
}

Candidate Question Pool:
{
Question ID: # Question ID
Difficulty: # Difficulty
Content: # Content
# IAMAG
Options: # Options
...
}

Available IDs: # List of Question ID
Output JSON format:
{
    "summary": "Summary the Statistics in history questions.Don't merely state the facts; instead, synthesize deeper, abstract, and even metaphysical patterns or principles.",
    "think": "Reasoning here",
    "question_index": "SELECTED_ID"
}
Only return the JSON object. DO NOT explain.

\end{lstlisting}
\end{tcolorbox}

\newpage

\newpage

\newpage
\medskip
{\small
\bibliography{main}
\bibliographystyle{plainnat}
}

\end{document}